\documentclass{article}

     \PassOptionsToPackage{numbers, compress}{natbib}

 \usepackage[preprint]{neurips_2019}

\usepackage[utf8]{inputenc} 
\usepackage[T1]{fontenc}    
\usepackage[hidelinks]{hyperref}       
\usepackage{url}            
\usepackage{booktabs}       
\usepackage{amsfonts}       
\usepackage{nicefrac}       
\usepackage{microtype}      

\usepackage{algorithm}
\usepackage{algorithmic}
\usepackage{amsmath,amssymb,amsthm}
\usepackage{graphicx}
\usepackage{subcaption}
\usepackage{multirow}
\usepackage{enumitem}

\theoremstyle{plain}

\newtheorem{defn}{Definition}

\newtheorem*{rep@theorem}{\rep@title}
\newcommand{\newreptheorem}[2]{%
	\newenvironment{rep#1}[1]{%
		\def\rep@title{#2 \ref{##1}}%
		\begin{rep@theorem}}%
		{\end{rep@theorem}}}
\newreptheorem{theorem}{Theorem}
\newreptheorem{lemma}{Lemma}
\newreptheorem{corollary}{Corollary}
\newreptheorem{prop}{Proposition}

\theoremstyle{definition}
\newtheorem{remark}{Remark}

\newcommand{\R}{\mathbb{R}}
\newcommand{\E}{\mathbb{E}}

\newcommand{\cS}{\mathcal{S}}

\newcommand{\cL}{\mathcal{L}}

\newcommand{\bw}{\boldsymbol{w}}

\newcommand{\bz}{\mathbf{z}}

\newcommand{\ssn}{\textsc{SSN}}
\newcommand{\asn}{\textsc{ASN}}
\newcommand{\mssn}{\textsc{MaxSSN}}
\newcommand{\Lmssn}{\mathcal{L}_{\mssn}}
\newcommand{\mssnadv}{\textsc{AdvMaxSSN}}
\newcommand{\Lmssnadv}{\mathcal{L}_{\mssnadv}}
\newcommand{\fdir}{f_{\text{direct}}}
\newcommand{\sgn}{\text{sgn}}
\newcommand{\APTD}{AP_{3D}}
\newcommand{\APBEV}{AP_{BEV}}
\newcommand{\numclean}{m_{\text{clean}}}
\newcommand{\numtune}{m_{\text{tune}}}
\newcommand{\iiter}{i_{\text{iter}}}
\newcommand{\thetafusion}{\boldsymbol{\theta}_{\text{fusion}}}
\newcommand{\dsum}{d_{\text{sum}}}

\title{On Single Source Robustness in Deep Fusion Models}

\author{ {\bf Taewan Kim}\thanks{This work was done when Taewan Kim was at the University of Texas at Austin, prior to joining Amazon.} \\
	The University of Texas at Austin\\
	Austin, TX\\
	\texttt{twankim@utexas.edu}
	\And
	{\bf Joydeep Ghosh}  \\
	The University of Texas at Austin\\
	Austin, TX\\
	\texttt{jghosh@utexas.edu}
}

\begin{document}

\maketitle

\begin{abstract}
  Algorithms that fuse multiple input sources benefit from both complementary and shared information. Shared information may provide robustness against faulty or noisy inputs, which is indispensable for safety-critical applications like self-driving cars. We investigate learning fusion algorithms that are robust against noise added to a single source. We first demonstrate that robustness against single source noise is not guaranteed in a linear fusion model. Motivated by this discovery, two possible approaches are proposed to increase robustness: a carefully designed loss with corresponding training algorithms for deep fusion models, and a simple convolutional fusion layer that has a structural advantage in dealing with noise. Experimental results show that both training algorithms and our fusion layer make a deep fusion-based 3D object detector robust against noise applied to a single source, while preserving the original performance on clean data.
\end{abstract}

\section{Introduction}\label{sec:intro}
Deep learning models have accomplished superior performance in several machine learning problems \citep{lecun2015deep} including object recognition \citep{krizhevsky2012imagenet,simonyan2014very,szegedy2015going,he2016deep,huang2017densely}, object detection \citep{ren2015faster,he2017mask,dai2016r,redmon2016you,liu2016ssd,redmon2017yolo9000} and speech recognition \citep{hinton2012deep,graves2013speech,sainath2013deep,chorowski2015attention,chan2016listen,chiu2018state}, which use either visual or audio sources. One natural way of improving a model's performance is to make use of multiple input sources relevant to a given task so that enough information can be extracted to build strong features. Therefore, deep fusion models have recently attracted considerable attention for autonomous driving \citep{kim2016robust,chen2017multi,qi2018frustum,ku2018joint}, medical imaging \citep{kiros2014stacked,wu2013online,simonovsky2016deep,liu2015multimodal}, and audio-visual speech recognition \citep{huang2013audio,mroueh2015deep,sui2015listening,chung2017lip}.

Two benefits are expected when fusion-based learning models are selected for a given problem. First, given adequate data, more information from multiple sources can enrich the model's feature space to achieve higher prediction performance, especially, when different input sources provide \textit{complementary information} to the model. This expectation coincides with a simple information theoretic fact: if we have multiple input sources $X_1,\cdots, X_{n_s}$ and a target variable $Y$, mutual information $I(;)$ obeys $I(Y;X_1,\cdots,X_{n_s}) \geq I(Y;X_i)~~(\forall i\in[n_s])$.

The second expected advantage is increased robustness against single source faults, which is the primary concern of our work. An underlying intuition comes from the fact that different sources may have \textit{shared information} so one sensor can partially compensate for others. This type of robustness is critical in real-world fusion models, because each source may be exposed to different types of corruption but not at the same time. For example, self-driving cars using an RGB camera and ranging sensors like LIDAR and radar are exposed to single source corruption. LIDARs and radars work fine at night whereas RGB cameras do not. Also, each source used in the model may have its own sensing device, and hence not necessarily be corrupted by some physical attack simultaneously with others. It would be ideal if the structure of machine learning based fusion models and shared information could compensate for the corruption and automatically guarantee robustness without additional steps.

This paper shows that a fusion model needs a supplementary strategy and a specialized structure to avoid vulnerability to noise or corruption on a single source. Our contributions are as follows:
\begin{itemize}
	\item We show that a fusion model learned with a standard robustness is not guaranteed to provide robustness against noise on a single source. Inspired by the analysis, a novel loss is proposed to achieve the desired robustness (Section \ref{sec:sinrobustness}).
	\item Two efficient training algorithms for minimizing our loss in deep fusion models are devised to ensure robustness without impacting performance on clean data (Section \ref{subsec:robust_algos}).
	\item We introduce a simple but an effective fusion layer which naturally reduces error by applying ensembling to latent convolutional features (Section \ref{subsec:feature_fusion}).
\end{itemize}
We apply our loss and the fusion layer to a complex deep fusion-based 3D object detector used in autonomous driving for further investigation in practice. Note that our findings can be easily generalized to other applications exhibiting intermittent defects in a subset of input sources, e.g., robustness given $k$ of $n_s$ corrupted sources, and single source robustness should be studied in depth prior to more general cases.

\section{Related Works}\label{sec:related}
Deep fusion models have been actively studied in object detection for autonomous vehicles. There exist two major streams classified according to their algorithmic structures: two-stage detectors with R-CNN (Region-based Convolutional Neural Networks) technique \citep{girshick2014rich,girshick2015fast,ren2015faster,dai2016r,he2017mask}, and single stage detectors for faster inference speed \citep{redmon2016you,redmon2017yolo9000,liu2016ssd}. 

Earlier deep fusion models extended Fast R-CNN \citep{girshick2015fast} to provide better quality of region proposals from multiple sources \citep{kim2016robust,braun2016pose}. With a high-resolution LIDAR, point cloud was used as a major source of the region proposal stage before the fusion step \citep{du2017car}, whereas F-PointNet \citep{qi2018frustum} used it for validating 2D proposals from RGB images and predicting 3D shape and location within the visual frustum. MV3D \citep{chen2017multi} extended the idea of region proposal network (RPN) \citep{ren2015faster} by generating proposals from RGB image, and LIDAR's front view and BEV (bird's eye view) maps. Recent works tried to remove region proposal stages for faster inference and directly fused LIDAR's front view depth image \citep{kim2018pedestrian} or BEV image \citep{wang2018fusing} with RGB images. ContFuse \citep{liang2018deep} utilizes both RGB and LIDAR's BEV images with a new continuous fusion scheme, which is further improved in MMF \citep{liang2019multi} by handling multiple tasks at once. Our experimental results are based on AVOD \citep{ku2018joint}, a recent open-sourced 3D object detector that generates region proposals from RPN using RGB and LIDAR's BEV images.

Compared to the active efforts in accomplishing higher performance on clean data, very few works have focused on robust learning methods in multi-source settings to the best of our knowledge. Adaptive fusion methods using gating networks weight the importance of each source automatically \citep{mees2016choosing,valada2017adapnet}, but these works lack in-depth studies of the robustness against single source faults. A recent work proposed a gated fusion at the feature level and applied data augmentation techniques with randomly chosen corruption methods \citep{kim2018robust}. In contrast, our training algorithms are surrogate minimization schemes for the proposed loss function, which is grounded from the analyses on underlying weakness of fusion methods. Also the fusion layer proposed in this paper focuses more on how to mix convolutional feature maps \textit{channel-wise} with simple trainable procedures. For extensive literature reviews, please refer to the recent survey papers about deep multi-modal learning methods in general \citep{ramachandram2017deep} and for autonomous driving \citep{feng2019deep}.


\section{Single Source Robustness of Fusion Models}\label{sec:sinrobustness}

\subsection{Regression on linear fusion data}\label{subsec:reg_model}
To show the vulnerability of naive fusion models, we introduce a simple data model and a fusion algorithm. Suppose $y$ is a linear function consisting of three different inherent (latent) components $z_i\in\R^{d_i}$ $(i\in\{1,2,3\})$. There are two input sources, $x_1$ and $x_2$. Here $\psi$'s are unknown functions.
\begin{equation}\label{eq:model_latent_rel}
y=\sum_{i=1}^3 {\beta_i} ^T z_i,~\text{ where }~z_1=\psi_1(x_1),~z_2=\psi_2(x_2),~z_3=\psi_{3,1}(x_1)=\psi_{3,2}(x_2)
\end{equation}
Our simple data model simulates a target variable $y$ relevant to two different sources, where each source has its own special information $z_1$ and $z_2$ and a shared one $z_3$. For example, if two sources are obtained from an RGB camera and a LIDAR sensor, one can imagine that any features related to objectness are captured in $z_3$ whereas colors and depth information may be located in $z_1$ and $z_2$, respectively. Our objective is to build a regression model by effectively incorporating information from the sources $(x_1,x_2)$ to predict the target variable $y$.

Now, consider a fairly simple setting $x_1=[z_1;z_3]\in\R^{d_1+d_3}$ and $x_2=[z_2;z_3]\in\R^{d_2+d_3}$, where $(\psi_1,\psi_2,\psi_{3,1},\psi_{3,2})$ can be defined  accordingly to satisfy (\ref{eq:model_latent_rel}). A straightforward fusion approach is to stack the sources, i.e. $x=[x_1;x_2]\in\R^{d_1+d_2+2d_3}$, and learn a linear model. Then, it is easy to show that there exists a feasible \textit{error-free model} for noise-free data:
\begin{equation}\label{eq:sol_direct}
\fdir(x_1,x_2)= h_1^T x_1 + h_2^T x_2 = (\beta_1^T z_1 + g_1^T z_3) + (\beta_2^T z_2 + g_2^T z_3),~s.t.~g_1+g_2=\beta_3
\end{equation}
where $h_1=[\beta_1;g_1],h_2=[\beta_2;g_2]$. Parameter vectors responsible for the shared information $z_3$ are denoted by $g_1$ and $g_2$.\footnote{In practice, $Y=[X_1,X_2]\begin{bmatrix}h_1\\h_2\end{bmatrix}$ has to be solved for $X_1\in \R^{n\times (d_1+d_3)},X_2\in\R^{n\times(d_2+d_3)}$ and $Y\in\R^n$ with enough number of $n$ data samples. Then a standard least squares solution using a pseudo-inverse gives $h_1=[\beta_1;\beta_3/2],~h_2=[\beta_2;\beta_3/2]$. This is equivalent to the solution robust against random noise added to all the sources at once, which is vulnerable to single source faults (Section \ref{subsec:siloss}).}


\paragraph{Unbalanced robustness (Motivation)}
Suppose the true parameters of data are scalar values, i.e. $\beta_i=c_i\in\R$ and influence of the complementary information is relatively small, $c_1\approx c_2$ and $c_3 \gg c_1$. Assume that the obtained error-free solution's parameters for $z_3$ are unbalanced, i.e. $g_1=\Delta$ and $g_2=c_3-\Delta$ with some weight parameter $\Delta\ll c_3$, so that $g_1$ gives a negligible contribution. Then add single source corruption $\delta_1=[\epsilon_1;\epsilon_3]$ and $\delta_2=[\epsilon_2;\epsilon_3]$ and compute absolute difference between the true data $y$ and the prediction from the corrupted data:
\begin{multline*}
|y-\fdir(x_1+\delta_1,x_2)|=|c_1\epsilon_1+\Delta\epsilon_3|,\hfill|y-\fdir(x_1,x_2+\delta_2)|=|c_2\epsilon_2+(c_3-\Delta)\epsilon_3|
\end{multline*}
In this case, adding noise to the source $x_2$ will give significant corruption to the prediction while $x_1$ is relatively robust because $|(c_3-\Delta)\epsilon_3|\gg |\Delta\epsilon_3|$ for any noise $\epsilon_3$ affecting $z_3$. This simple example illustrates that additional training strategies or components are indispensable to achieve robust fusion model working even if one of the sources is disturbed. The next section introduces a novel loss for a balanced robustness against a fault in a single source.

\subsection{Robust learning for single source noise}\label{subsec:siloss}
Fusion methods are not guaranteed to provide robustness against faults in a single source without additional supervision. Also, we demonstrate that naive regularization or robust learning methods are not sufficient for the robustness later in this section. Therefore, a supplementary constraint or strategy needs to be considered in training which can correctly guide learning parameters for the desired robustness.

One essential requirement of fusion models is showing \textit{balanced performance} regardless of corruption added to any source. If the model is significantly vulnerable to corruption in one source, this model becomes untrustworthy and we need to balance the degradation levels of different input sources' faults. For example, suppose there is a model robust against noise in RGB channels but shows huge degradation in performance for any fault of LIDAR. Then the overall system should be considered untrustworthy, because there exist certain corruption or environments which can consistently fool the model. Our loss, \mssn~(Maximum Single Source Noise), for such robustness is introduced to handle this issue and further analyses are provided under the linear fusion data model explained in Section \ref{subsec:reg_model}. This loss makes the model focus more on corruption of a single source, \ssn, rather than focusing on noise added to all the sources at once, \asn.

\begin{defn}\label{dfn:maxssn_loss}
	For multiple sources $x_1,\cdots,x_{n_s}$ and a target variable $y$, denote a predefined loss function by $\cL$. If each source $x_i$ is perturbed with some additive noise $\epsilon_i$ for $i\in[n_s]$, \mssn~loss for a model $f$ is defined as follows:
	\begin{align*}
	\Lmssn(f,\epsilon)\triangleq \max_{i} \left\{ \cL\left( y,f(x_1,\cdots,x_{i-1},x_i+\epsilon_i,x_{i+1},\cdots,x_{n_s}) \right)\right\}_{i=1}^{n_s}
	\end{align*}
\end{defn}

Another key principle in our robust training is to \textit{retain the model's performance on clean data}. Although techniques like data augmentation help improving a model's generalization error in general, learning a model robust against certain types perturbation including adversarial attacks may harm the model's accuracy on non-corrupt data \citep{tsipras2018robustness}. Deterioration in the model's ability on normal data is an unwanted side effect, and hence our approach aims to avoid this.

\paragraph{Random noise}\label{par:rand_noise}
To investigate the importance of our \mssn~loss, we revisit the linear fusion data model with the optimal direct fusion model $\fdir$ of the regression problem introduced in Section \ref{subsec:reg_model}. Suppose the objective is to find a model with robustness against single source noises, while preserving error-free performance, i.e., unchanged loss under clean data. For the noise model, consider $\epsilon=[\delta_1;\delta_2]$ where $\delta_1=[\epsilon_1;\epsilon_3]$ and $\delta_2=[\epsilon_2;\epsilon_4]$, which satisfy $\E[\epsilon_i]=0$, $Var(\epsilon_i)=\sigma^2 I$, and $\E[\epsilon_i\epsilon_j^T]=0$ for $i\neq j$. Note that noises added to the shared information, $\epsilon_3$ and $\epsilon_4$, are not identical, which resembles direct perturbation to the input sources in practice. For example, noise directly affecting a camera lens does not need to perturb other sources.

\paragraph{Optimal fusion model for \mssn} The robust linear fusion model $f(x_1,x_2)=(w_1^T z_1+g_1^T z_3) + (w_2^T z_1+g_2^T z_3)$ is found by minimizing $\cL_\mssn(f,\epsilon)$ over parameters $w_1,w_2,g_1$ and $g_2$. As shown in the previous section, any $\fdir$ satisfying $w_1=\beta_1,w_2=\beta_2$ and $g_1+g_2=\beta_3$ should achieve zero-error. Therefore, overall optimization problem can be reduced to the following one:
\begin{equation}\label{eq:opt_rand}
\min_{g_1,g_2}\max\left\{\cL\left(y,\fdir(x_1+\delta_1,x_2)\right) ,\cL\left(y,\fdir(x_1,x_2+\delta_2)\right)\right\}\hspace{1em}s.t.~~g_1+g_2=\beta_3
\end{equation}
If we use a standard expected squared loss $\cL(y,f(x_1,x_2))=\E[(y-f(x_1,x_2))^2]$ and solve the optimization problem, the following solution $\cL_{\mssn}^*$ with corresponding parameters $g_1^*,g_2^*$ can be obtained, and there exist three cases based on the relative sizes of $||\beta_i||_2$'s.
\begin{equation}\label{eq:sol_linear_rand}
\left(\cL_{\mssn}^*,~g_1^*,~g_2^*\right)=\begin{cases}
\left(\sigma^2||\beta_2||_2^2,~\beta_3,~0 \right) \hfill\text{if } ||\beta_1||_2^2+||\beta_3||_2^2\leq ||\beta_2||_2^2\\
\left(\sigma^2||\beta_1||_2^2,~0,~\beta_3 \right) \hfill\text{if } ||\beta_2||_2^2+||\beta_3||_2^2\leq ||\beta_1||_2^2\\
\left(\sigma^2\left(\frac{||\beta_1||_2^2+||\beta_2||_2^2}{2}+\frac{||\beta_3||_2^2}{4}+\frac{(||\beta_2||_2^2-||\beta_1||_2^2)^2}{4||\beta_3||_2^2}\right), \right.\\
\hspace{.5em}\left.\frac{1}{2}\left(1+\frac{||\beta_2||_2^2-||\beta_1||_2^2}{||\beta_3||_2^2}\right),~\frac{1}{2}\left(1-\frac{||\beta_2||_2^2-||\beta_1||_2^2}{||\beta_3||_2^2}\right) \right)  \hspace{2em}\text{otherwise}
\end{cases}
\end{equation}
The three cases reflect the relative influence of each weight vector for $z_i$. For instance, if $z_2$ has larger importance compared to the rest in generating $y$, the optimal way of balancing the effect of noise over $z_3$ is to remove all the influence of $z_2$ in $x_2$ by setting $g_2=0$. When neither of $z_1$ nor $z_2$ dominates the importance, i.e. $\left|\frac{||\beta_2||_2^2-||\beta_1||_2^2}{||\beta_3||_2^2}\right| < 1$, the optimal solution tries to make $\cL\left(y,\fdir(x_1+\delta_1,x_2)\right)= \cL\left(y,\fdir(x_1,x_2+\delta_2)\right)$.

\paragraph{Comparison with the standard robust fusion model} Minimizing loss with noise added to a model's input is a standard process in robust learning. The same strategy can be applied to learn fusion models by considering all sources as a single combined source, then add noise to all the sources at once. However, this simple strategy cannot achieve low error in terms of the single source robustness. The optimal solution to $\min_{g_1,g_2} \E[(y-\fdir(x_1+\delta_1,x_2+\delta_2))^2]$, a least squares solution, is achieved when $g_1=g_2=\frac{\beta_3}{2}$. The corresponding \mssn~loss can be evaluated as $\Lmssn'=\sigma^2\max\left\{||\beta_1||_2^2+\frac{1}{4}||\beta_3||_2^2,||\beta_2||_2^2+\frac{1}{4}||\beta_3||_2^2 \right\}$. A nontrivial gap exists between $\Lmssn$ and $\Lmssn'$, which is directly proportional to the data model's inherent characteristics:
\begin{equation}\label{eq:gap_linear_rand}
\Lmssn'-\Lmssn^*\geq\begin{cases}
\frac{1}{4}||\beta_3||_2^2 & \text{if } \left|\frac{||\beta_2||_2^2-||\beta_1||_2^2}{||\beta_3||_2^2}\right| \geq 1\\
\frac{1}{4}\left| ||\beta_2||_2^2-||\beta_1||_2^2 \right| & \text{otherwise}
\end{cases}
\end{equation}
If either $z_1$ or $z_2$ has more influence on the target value $y$ than the other components, single source robustness of the model trained by \mssn~loss is better than the fusion model for the general noise robustness with an amount proportional to the influence of shared feature $z_3$. Otherwise, the gap's lower bound is proportional to the difference in complementary information, $|||\beta_2||_2^2-||\beta_1||_2^2|/4$.


\begin{remark}
	In linear systems such as the one studied above, having redundant information in the feature space is similar to multicollinearity in statistics. In this case, feature selection methods usually try to remove such redundancy. However, this redundant or shared information helps preventing degradation of the fusion model when a subset of the input sources are corrupted. 
\end{remark}
\begin{remark}
	Similar analyses and a loss definition against \textit{adversarial attacks} \citep{goodfellowexplaining} are provided in appendix \ref{subsec:adv_attack}.
\end{remark} 


\section{Robust Deep Fusion Models}\label{sec:robust_deep}
In simple linear settings, our analyses illustrate that using $\mssn$~loss can effectively minimize the degradation of a fusion model's performance. This suggests a training strategy for complex deep fusion models to be equipped with robustness against single source faults. A principal factor considered in designing a common framework for our algorithms is the preservation of model's performance on clean data while minimizing a loss for defending corruption. Therefore, our training algorithms use \textit{data augmentation} to encounter both clean and corrupted data. The second way of achieving robustness is to take advantage of the fusion method's structure. A simple but effective method of mixing convolutional features coming from different input sources is introduced later in this section.

\subsection{Robust training algorithms for single source noise}\label{subsec:robust_algos}
In the previous section, we solve problem (\ref{eq:opt_rand}) by optimizing over flexible parameters $g_1$ and $g_2$. If the parts of input sources contributing to $z_3$ are known, then indeed we can achieve this goal. In practice however, it is difficult to know which parts of an input source (or latent representation) are related to shared information and which parameters are flexible. Therefore, our common training framework alternately provides \textit{clean samples} and \textit{corrupted samples} per iteration to preserve the original performance of the model on uncontaminated data.\footnote{We also try \textit{fine-tuning} only a subset of the model's parameters, $\thetafusion\subset f$, to preserve essential parts for extracting features from normal data. Although this strategy is similar to optimizing over only $g_1$ and $g_2$ in our linear fusion case, training the whole network from the beginning shows better performance in practice. See Appendix \ref{sec:append_exp} for a detailed comparison.} On top of this strategy, one standard robust training scheme and two algorithms for minimizing $\mssn$~ loss are introduced for handling robustness against noise in different sources.

\paragraph{Standard robust training method} A standard robust training algorithm can be developed by considering all $n_s$ sources as a single combined source. Given noise generating functions $\varphi_i(\cdot)$ ($i\in[n_s]$), the algorithm generates and adds corruption to all the sensors at once. Then the corresponding loss can be computed to update parameters using back-propagation. This algorithm is denoted by \textsc{TrainASN} and tested in experiments to investigate whether the procedure is also able to cover robustness against single source noise.

\begin{figure}[th]
	\begin{minipage}[t]{.48\linewidth}
		\begin{algorithm} [H]
			\caption{\textsc{TrainSSN}}
			\label{alg:TrainSSN}
			\begin{algorithmic} 
				\FOR{$i_{\text{iter}}=1$ to $m$}
				\STATE Sample $(y,\{x_i\}_{i=1}^{n_s})$
				\IF{$\iiter\equiv 1$ (mod $2$)}
				\FOR{$j=1$ to $n_s$}
				\STATE Generate noise $\epsilon_j=\varphi_j(x_j)$
				\STATE $\hat{\cL}^{(i_\text{iter})}_j\leftarrow\cL(y,f(\{x_j+\epsilon_j,x_{-j}\}))$\\
				\ENDFOR
				\STATE $\cL^{(i_\text{iter})}\leftarrow\max_{j}\hat{\cL}^{(i_\text{iter})}_j$
				\ELSE
				\STATE $\cL^{(i_\text{iter})}\leftarrow\cL(y,f(\{x_i\}_{i=1}^{n_s}))$
				\ENDIF
				\STATE Update $f$ using $\nabla\cL^{(i_\text{iter})}$
				\ENDFOR
			\end{algorithmic}
		\end{algorithm}
	\end{minipage}
	\hfill
	\begin{minipage}[t]{.48\linewidth}
		\begin{algorithm} [H]
			\caption{\textsc{TrainSSNAlt}}
			\label{alg:TrainSSNAlt}
			\begin{algorithmic} 
				\FOR{$i_{\text{iter}}=1$ to $m$}
				\STATE Sample $(y,\{x_i\}_{i=1}^{n_s})$
				\IF{$\iiter\equiv 1$ (mod 2)}
				\STATE $j\leftarrow (\lfloor\iiter/2\rfloor \text{ mod }n_s)+1$
				\STATE Generate noise $\epsilon_j=\varphi_j(x_j)$
				\STATE $\cL^{(i_\text{iter})}\leftarrow\cL(y,f(\{x_j+\epsilon_j,x_{-j}\}))$
				\ELSE
				\STATE $\cL^{(i_\text{iter})}\leftarrow\cL(y,f(\{x_i\}_{i=1}^{n_s}))$
				\ENDIF
				\STATE Update $f$ using $\nabla\cL^{(i_\text{iter})}$
				\ENDFOR
			\end{algorithmic}
		\end{algorithm}
	\end{minipage}
\end{figure}

\paragraph{Minimization of \mssn~loss} 
Minimization of the \mssn~loss requires $n_s$ (number of input sources) forward-propagations within one iteration. Each propagation needs a different set of corrupted samples generated by adding single source noise to the fixed clean mini-batch of data. There are two possible approaches to compute gradients properly from these multiple passes. First, we can run back-propagation $n_s$ times to save the gradients temporarily without updating any parameters, then the saved gradients with the maximum loss is used for updating parameters. However, this process requires not only $n_s$ forward and backward passes but also large memory usage proportional to $n_s$ for saving the gradients. Another reasonable approach is to run $n_s$ forward passes to find the maximum loss and compute gradients by going back to the corresponding set of corrupted samples. Algorithm \ref{alg:TrainSSN} adopts this idea for its efficiency, $n_s+1$ forward passes and one back-propagation. A faster version of the algorithm, \textsc{TrainSSNAlt}, is also considered since multiple forward passes may take longer as the number of sources increases. This algorithm ignores the maximum loss and alternately augments corrupted data. By a slight abuse of notation, symbols used in our algorithms also represent the iteration steps with the size of mini-batches greater than one. Also, $f(x_1,\cdots,x_{j-1},x_j+\epsilon_j,x_{j+1},\cdots,x_{n_s})$ is shortened to $f(\{x_j+\epsilon_j,x_{-j}\})$ in the algorithms.

\subsection{Feature fusion methods}\label{subsec:feature_fusion}
Fusion of features extracted from multiple input sources can be done in various ways \citep{chen2017multi}. One of the popular methods is to fuse via an element-wise mean operation \citep{ku2018joint}, but this assumes that each feature must have a same shape, i.e., width, height, and number of channels for a 3D feature. An element-wise mean can be also viewed as averaging channels from different 3D features, and it has an underlying assumption that the channels of each feature should share same information regardless of the input source origin. Therefore, the risk of becoming vulnerable to single source corruption may increase with this simple mean fusion method.

\begin{figure}[ht]
	\centering
	\includegraphics[width=.95\linewidth]{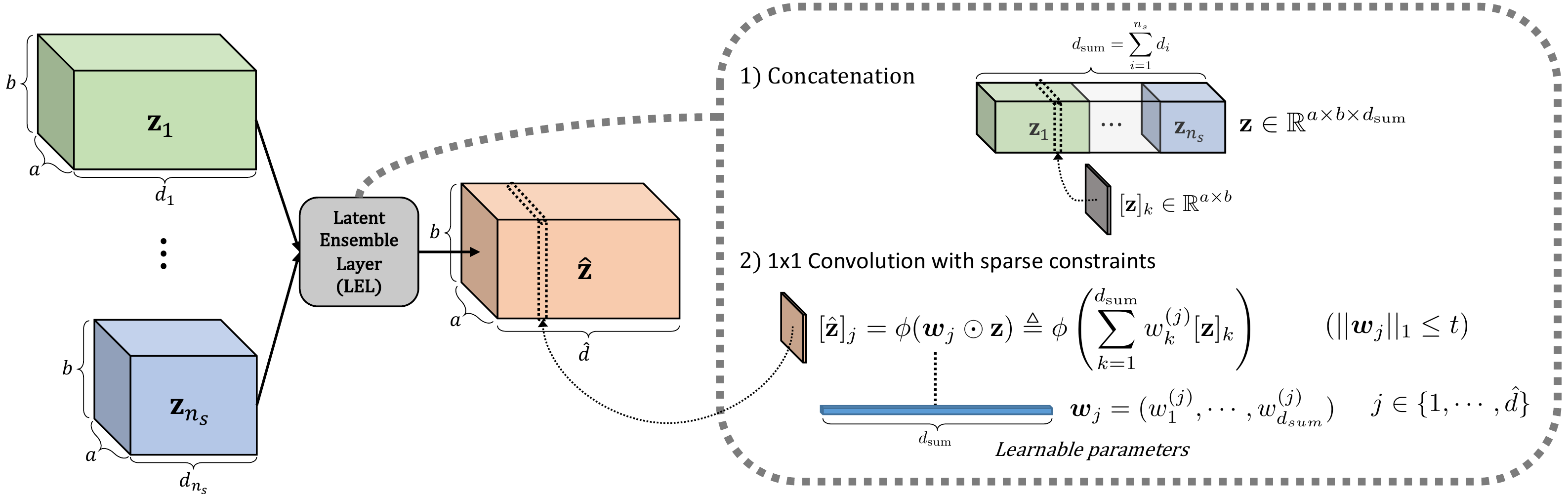}
	\caption{Latent ensemble layer (LEL)}
	\label{fig:lel}
\end{figure}
Our fusion method, latent ensemble layer (LEL), is devised for three objectives: (i) maintaining the known advantage---error reduction---of ensemble methods \citep{tumer1996error,tumer1996analysis}, (ii) admitting source-specific features to survive even after the fusion procedure, and (iii) allowing each source to provide a different number of channels. The proposed layer learns parameters so that channels of the 3D features from the different sources can be selectively mixed. Sparse constraints are introduced to let the training procedure find good subsets of channels to be fused across the $n_s$ feature maps. For example, mixing the $i^{\text{th}}$ channel of the convolutional feature from an RGB image with the $j^{\text{th}}$ and $k^{\text{th}}$ channels of the LIDAR's latent feature is possible in our LEL, whereas in an element-wise mean layer the $i^{\text{th}}$ latent channel from RGB is only mixed with the other sources' $i^{\text{th}}$ channels.

In practice, this layer can be easily constructed by using $1\times 1$ convolutions with the ReLU activation and $\ell_1$ constraints. We also apply an activation function to supplement a semi-adaptive behavior to the fusion procedure. Depth of the output channel is set to $\hat{d}=\max_i\{d_i\}$ and we set the hyper-parameter for $\ell_1$ constraint as 0.01 in the experiments. Definition \ref{dfn:lel} explains the details of our LEL, and Figure \ref{fig:lel} visualizes the overall process. 

\begin{defn}[Latent ensemble layer]\label{dfn:lel}
	Suppose we have $n_s$ convolutional features $\bz_i\in\R^{a\times b \times d_i}$ from different input sources $(i\in[n_s])$, which can be stacked as $\bz=(\bz_1,\cdots,\bz_m) \in \R^{a \times b \times \dsum}$ $(\dsum=\sum_{i=1}^m d_i)$. The $k^{th}$ channel of the stacked feature is denoted by $[\bz]_k\in\R^{a\times b}$. Let $\bw_j=(w_1^{(j)},\cdots,w_{\dsum}^{(j)})$ be a $\dsum$-dimensional weight vector to mix $\bz_i$'s in channel-wise fashion. Then LEL outputs $\hat{\bz}\in\R^{a\times b \times \hat{d}}$ where each channel is computed as $[\hat{\bz}]_j=\phi(\bw_j\odot\bz)\triangleq\phi\left(\sum_{k=1}^{\dsum}w_{k}^{(j)}[\bz]_k\right)$, with some activation function $\phi$ and sparse constraints $||\bw_j||_0\leq t$ for all $j \in \{1,\cdots,\hat{d}\}$. 
\end{defn}

\section{Experimental Results}\label{sec:exp_results}
We test our algorithms and the LEL fusion method on 3D and BEV object detection tasks using the car class of the KITTI dataset \citep{geiger2012we}. 3D detection is both an important problem in self-driving cars and one where multiple sensors can contribute fruitfully by providing both complementary and shared information. In contrast, models for 2D object detection heavily rely on RGB data, which typically dominates other modalities. As our experiments include random generation of corruption, each task is evaluated 5 times to compare average scores (reported with 95\% confidence intervals), and thus a validation set is used for ease of manipulating data and repetitive evaluation. We follow the split of \citet{ku2018joint}, 3712 and 3769 frames for training and validation sets, respectively. Results are reported based on three difficulty levels defined by KITTI (easy, medium, hard) and a standard metric Average Precision (AP) is used. A recent open-sourced 3D object detector AVOD \citep{ku2018joint} with a feature pyramid network is selected as a baseline algorithm. 

Four different algorithms are compared: AVOD trained on (i) clean data, (ii) data augmented with ASN samples (\textsc{TrainASN}), (iii) SSN augmented data with direct \mssn~loss minimization (\textsc{TrainSSN}), and (iv) SSN augmented data (\textsc{TrainSSNAlt}). The AVOD architecture is varied to use either element-wise mean fusion layers or our LELs. We follow the original training setups of AVOD, e.g., 120k iterations using an ADAM optimizer with an initial learning rate of 0.0001.\footnote{Our methods are implemented with TensorFlow on top of the official AVOD code. The computing machine has a Intel Xeon E5-1660v3 CPU with Nvidia Titan X Pascal GPUs. The source code is available at \url{https://github.com/twankim/avod_ssn}.}


\begin{figure}[t]
	\centering
	\begin{subfigure}{\linewidth}
		\centering
		\includegraphics[width=.49\linewidth]{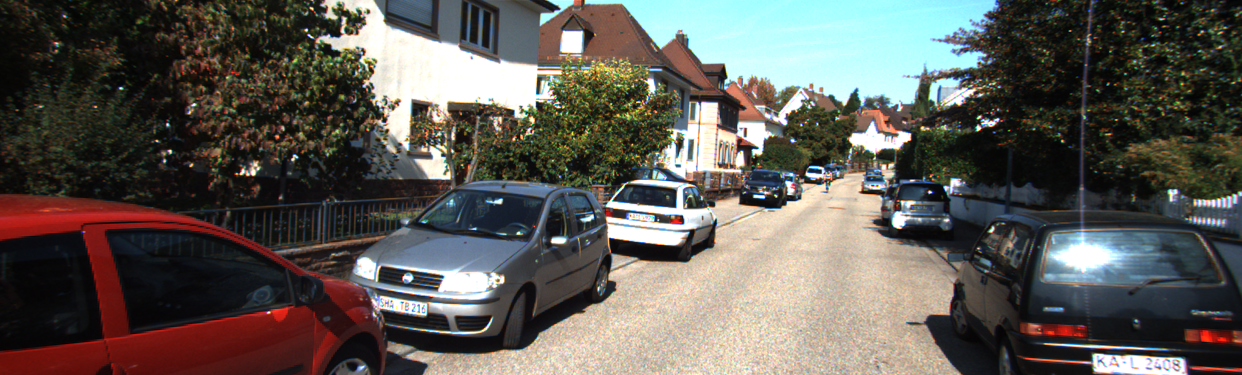}
		\hfill
		\includegraphics[width=.49\linewidth]{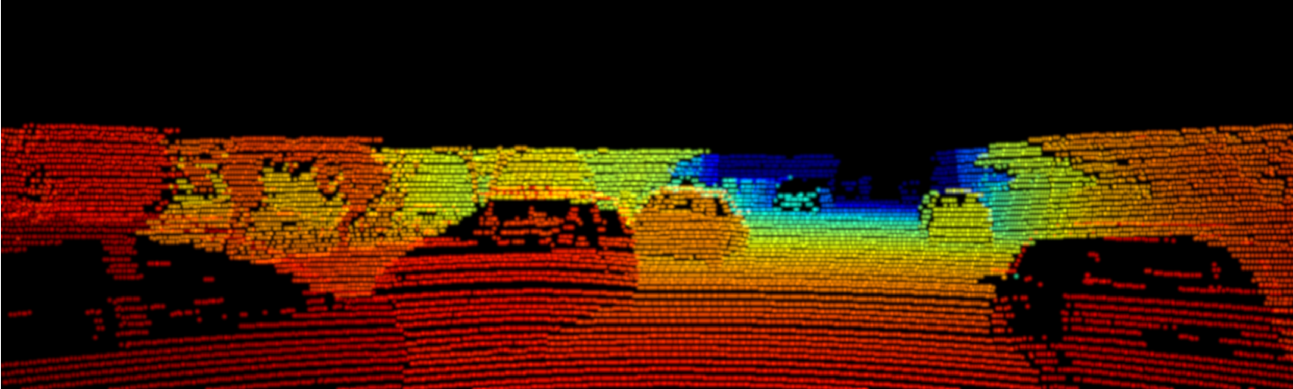}
		\caption{Original}
	\end{subfigure}
	\\
	\vspace{.3em}
	\begin{subfigure}{\linewidth}
		\centering
		\includegraphics[width=.49\linewidth]{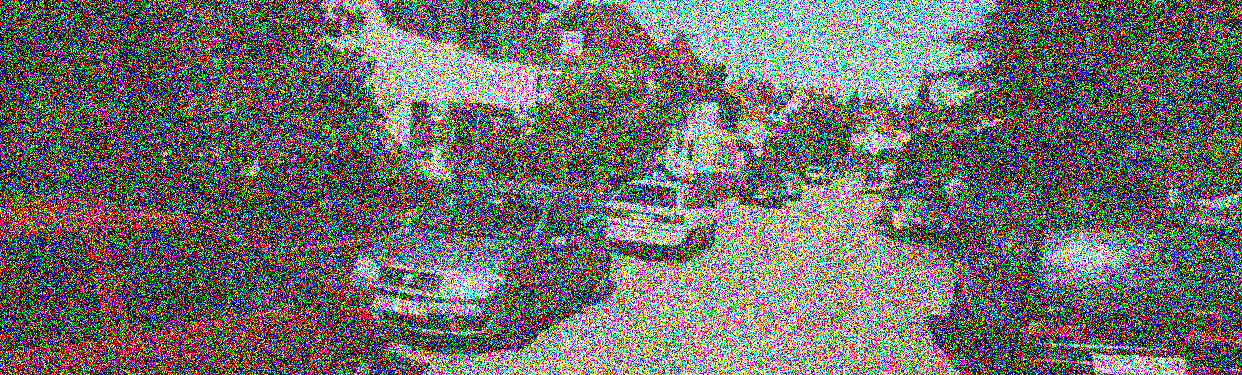}
		\hfill
		\includegraphics[width=.49\linewidth]{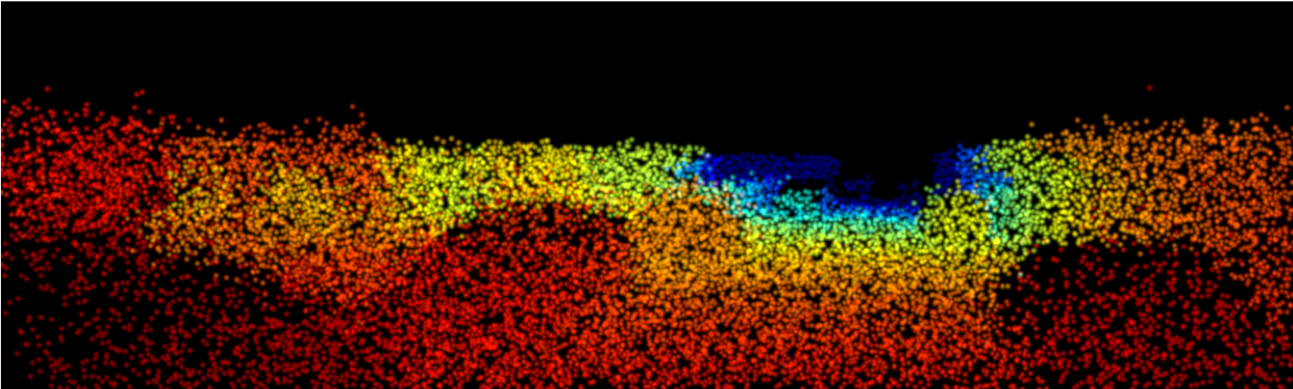}
		\caption{Gaussian noise}
	\end{subfigure}
	\\
	\vspace{.3em}
	\begin{subfigure}{\linewidth}
		\centering
		\includegraphics[width=.49\linewidth]{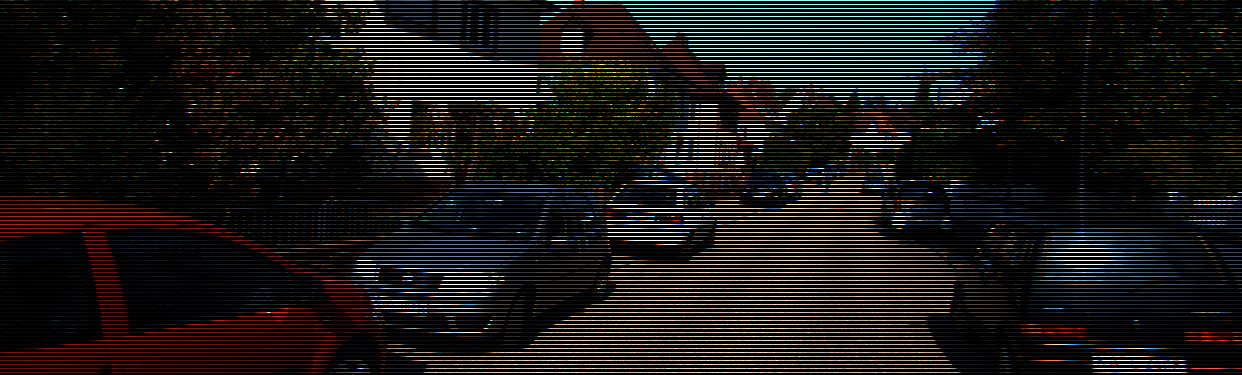}
		\hfill
		\includegraphics[width=.49\linewidth]{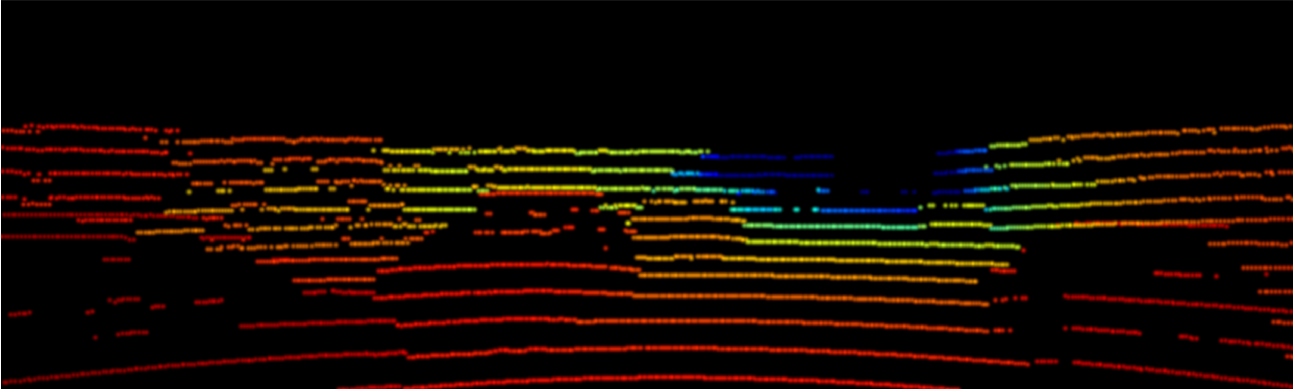}
		\caption{Downsampling}
	\end{subfigure}
	\caption{Visualization of corrupted samples, \textit{(left)} RGB images \textit{(right)} LIDAR point clouds. The point clouds are projected onto the 2D image plane for easier visual comparison.}
	\label{fig:ex_samples}
\end{figure}

\paragraph{Corruption methods} \textit{Gaussian noise} generated i.i.d. with $\mathcal{N}(0,\sigma_{\text{Gaussian}}^2)$ is directly added to the pixel value of an image ($r,g,b$) and the coordinate value of a LIDAR's point ($x,y,z$). $\sigma_{\text{Gaussian}}$ is set to $0.75\tau$ experimentally with $\tau_{\text{RGB}}=255$ and $\tau_{\text{LIDAR}}=0.2$. The second method \textit{downsampling} selects only 16 out of 64 lasers of LIDAR data. To match this effect, 3 out of 4 horizontal lines of an RGB image are deleted. Effects of corruption on each input source are visualized in Figure \ref{fig:ex_samples}, where the color of a 2D LIDAR image represents a distance from the sensor. Although our analyses in Section \ref{subsec:siloss} assume the noise variances to be identical, it is nontrivial to set equal noise levels for different modalities in practice, e.g., RGB pixels vs. points in a 3D space. Nevertheless, an underlying objective of our \mssn~loss, balancing the degradation rates of different input sources' faults, does not depend on the choice of noise types or levels.

\paragraph{Evaluation metrics for single source robustness} To assess the robustness against single source noise, a new metric minAP is introduced. The AP score is evaluated on the dataset with a single corrupted input source, then after going over all $n_s$ sources, minAP reports the lowest score among the $n_s$ AP scores. Our second metric maxDiffAP computes the maximum absolute difference among the scores, which measures the balance of different input sources' single source robustness; low value of maxDiffAP means the well-balanced robustness.

\begin{table}[th]
	\caption{Car detection (3D/BEV) performance of AVOD with \textit{element-wise mean} fusion layers and \textit{latent ensemble layers (LEL)} against \textit{Gaussian} SSN on the KITTI validation set.}
	\label{table:result_car_simple_lel_rand_ssn_main}
	\begin{center}
		\begin{scriptsize}
			\begin{tabular}{lcccccc}
				\toprule
				(Data) Train Algo. & Easy & Moderate & Hard & Easy & Moderate & Hard\\
				\midrule
				\multicolumn{7}{c}{\textit{\textbf{Fusion method}: Mean}}\vspace{.3em}\\
				(Clean Data) & \multicolumn{3}{c}{$\APTD (\%)$} & \multicolumn{3}{c}{$\APBEV (\%)$}\\ 
				\cmidrule(r){2-4} \cmidrule(r){5-7}
				AVOD \citep{ku2018joint} & \bf{76.41} & \bf{72.74} & \bf{66.86} & 89.33 & 86.49 & \bf{79.44}\\
				+\textsc{TrainASN} & 75.96 & 66.68 & 65.97 & 88.63 & 79.45 & 78.79\\
				+\textsc{TrainSSN} & 76.28 & 67.10 & 66.51 & 88.86 & 79.60 & 79.11\\
				+\textsc{TrainSSNAlt} & \textbf{77.46} & 67.61 & 66.06 & \bf{89.68} & \bf{86.71} & 79.41\\
				\midrule
				(Gaussian SSN) & \multicolumn{3}{c}{$\min\APTD (\%)$} & \multicolumn{3}{c}{$\min\APBEV (\%)$}\\
				\cmidrule(r){2-4} \cmidrule(r){5-7}
				AVOD \citep{ku2018joint} & $47.41 {\pm} 0.28$ & $41.84 {\pm} 0.17$ & $36.47 {\pm} 0.16$ & $65.63 {\pm} 0.28$ & $58.02 {\pm} 0.23$ & $50.43 {\pm} 0.14$\\
				+\textsc{TrainASN} & $61.53 {\pm} 0.57$ & $52.72 {\pm} 0.08$ & $47.25 {\pm} 0.13$ & $87.71 {\pm} 0.14 $ & $78.37 {\pm} 0.06$ & $77.85 {\pm} 0.08$ \\
				+\textsc{TrainSSN} & $71.65 {\pm} 0.31$ & $\mathbf{62.14 {\pm} 0.08}$ & $\mathbf{56.78 {\pm} 0.12}$ & $88.21 {\pm} 0.08$ & $78.90 {\pm} 0.09$ & $\mathbf{77.92 {\pm} 0.11}$\\
				+\textsc{TrainSSNAlt} & $\mathbf{71.66 {\pm} 0.48}$ & $57.61 {\pm} 0.12$ & $55.90 {\pm} 0.11$
				& $\mathbf{89.42 {\pm} 0.04}$ & $\mathbf{79.56 {\pm} 0.06}$ & $\mathbf{77.92 {\pm} 0.05}$\\
				\midrule
				(Gaussian SSN) & \multicolumn{3}{c}{$\max \text{Diff}\APTD (\%)$} & \multicolumn{3}{c}{$\max \text{Diff}\APBEV (\%)$}\\
				AVOD \citep{ku2018joint} & $26.70 {\pm} 0.52$ & $22.42 {\pm} 0.29$ & $20.92 {\pm} 0.25$ & $22.27 {\pm} 0.41$ & $20.76 {\pm} 0.33$ & $20.09 {\pm} 0.20$\\
				+\textsc{TrainASN} & $14.48 {\pm} 0.82$ & $12.72 {\pm} 0.33$ & $11.18 {\pm} 0.27$ & $0.88 {\pm} 0.22$ & $0.48 {\pm} 0.13$ & $0.28 {\pm} 0.12$ \\
				+\textsc{TrainSSN} & $\mathbf{3.71 {\pm} 0.46}$ & $\mathbf{3.42 {\pm} 0.25}$ & $7.50 {\pm} 0.25$ & $0.36 {\pm} 0.17$ & $\mathbf{0.04 {\pm} 0.15}$ & $0.71 {\pm} 0.17$\\
				+\textsc{TrainSSNAlt} & $5.55 {\pm} 0.81$ & $8.73 {\pm} 0.32$ & $\mathbf{2.91 {\pm} 0.22}$ & $\mathbf{0.09 {\pm} 0.14}$ & $0.13 {\pm} 0.11$ & $\mathbf{0.18 {\pm} 0.11}$\\
				\midrule
				\multicolumn{7}{c}{\textit{\textbf{Fusion method}: Latent Ensemble Layer}}\vspace{.3em}\\
				(Clean Data) & \multicolumn{3}{c}{$\APTD (\%)$} & \multicolumn{3}{c}{$\APBEV (\%)$}\\ 
				\cmidrule(r){2-4} \cmidrule(r){5-7}
				AVOD \citep{ku2018joint} & \bf{77.79} & \bf{67.69} & \bf{66.31} & \bf{88.90} & \bf{85.64} & \bf{78.86}\\
				+\textsc{TrainASN} & 75.00 & 64.75 & 58.28 & 88.30 & 78.60 & 77.23\\
				+\textsc{TrainSSN} & 74.25 & 65.00 & 63.83 & 87.88 & 78.84 & 77.66\\
				+\textsc{TrainSSNAlt} & 76.04 & 66.42 & 64.41 & 88.80 & 79.53 & 78.53\\
				\midrule
				(Gaussian SSN) & \multicolumn{3}{c}{$\min\APTD (\%)$} & \multicolumn{3}{c}{$\min\APBEV (\%)$}\\
				\cmidrule(r){2-4} \cmidrule(r){5-7}
				AVOD \citep{ku2018joint} & $61.97 {\pm} 0.55$ & $53.95 {\pm} 0.42$ & $47.24 {\pm} 0.27$ & $79.44 {\pm} 0.09$ & $72.46 {\pm} 3.14$ & $68.25 {\pm} 0.06$\\
				+\textsc{TrainASN} & $\mathbf{74.24 {\pm} 0.38}$ & $58.25 {\pm} 0.16$ & $\mathbf{56.13 {\pm} 0.10}$ & $88.10 {\pm} 0.26$ & $\mathbf{78.19 {\pm} 0.13}$ & $70.42 {\pm} 0.07$ \\
				+\textsc{TrainSSN} & $68.16 {\pm} 0.88$ & $\mathbf{60.39 {\pm} 0.38}$ & $56.04 {\pm} 0.28$ & $\mathbf{88.12 {\pm} 0.16}$ & $78.17 {\pm} 0.06$ & $70.21 {\pm} 0.05$\\
				+\textsc{TrainSSNAlt} & $68.63 {\pm} 0.40$ & $55.48 {\pm} 0.16$ & $54.42 {\pm} 0.17$
				& $86.51 {\pm} 0.46$ & $76.85 {\pm} 0.11$ & $\mathbf{71.95 {\pm} 2.72}$\\
				\bottomrule
			\end{tabular}
		\end{scriptsize}
	\end{center}
\end{table}

\begin{table}[th]
	\caption{Car detection (3D/BEV) performance of AVOD with \textit{latent ensemble layers (LEL)} against \textit{downsampling} \ssn~on the KITTI validation set.}
	\label{table:result_car_lel_full_asn_ssn_downsample_main}
	\begin{center}
		\begin{scriptsize}
			\begin{tabular}{lcccccc}
				\toprule
				(Data) Train Algo. & Easy & Moderate & Hard & Easy & Moderate & Hard\\
				\midrule
				(Clean Data) & \multicolumn{3}{c}{$\APTD (\%)$} & \multicolumn{3}{c}{$\APBEV (\%)$}\\ 
				\cmidrule(r){2-4} \cmidrule(r){5-7}
				AVOD \citep{ku2018joint} & \bf{77.79} & \bf{67.69} & \bf{66.31} & 88.90 & \bf{85.64} & \bf{78.86}\\
				+\textsc{TrainASN} & 71.74 & 61.78 & 60.26 & 87.29 & 77.08 & 75.89\\
				+\textsc{TrainSSN} & 75.54 & 66.26 & 63.72 & 88.07 & 79.18 & 78.03\\
				+\textsc{TrainSSNAlt} & 76.22 & 66.05 & 63.87 & \bf{89.00} & 79.65 & 78.03\\
				\midrule
				(Downsample SSN) & \multicolumn{3}{c}{$\min\APTD (\%)$} & \multicolumn{3}{c}{$\min\APBEV (\%)$}\\
				\cmidrule(r){2-4} \cmidrule(r){5-7}
				AVOD \citep{ku2018joint} & 61.70 & 51.66 & 46.17 & 86.08 & 69.99 & 61.55\\
				+\textsc{TrainASN} & 65.74 & 53.49 & 51.35 & 82.27 & 67.88 & 65.79\\
				+\textsc{TrainSSN} & \bf{73.33} & \bf{57.85} & \bf{54.91} & \bf{86.61} & \bf{76.07} & \bf{68.59}\\
				+\textsc{TrainSSNAlt} & 64.77 & 53.34 & 48.29 & 85.27 & 69.87 & 67.77\\
				\bottomrule
			\end{tabular}
		\end{scriptsize}
	\end{center}
\end{table}

\paragraph{Results} 
When the fusion model uses the element-wise mean fusion (Table \ref{table:result_car_simple_lel_rand_ssn_main}), \textsc{TrainSSN} algorithm shows the best single source robustness against Gaussian SSN while preserving the original performance on clean data (only small decrease in the moderate BEV detection)\footnote{In practice, it is difficult to identify flexible parameters related to shared information in advance, and also the design goal becomes a soft rather than a hard constraint. Therefore there is minor degradation in performance, to pay for the added robustness.}. Also a balance of the both input sources' performance is dramatically decreased compared to the models trained without robust learning and a naive \textsc{TrainASN} method.

Encouragingly, AVOD model constructed with our LEL method already achieves relatively high robustness without any robust learning strategies compared to the mean fusion layers. For all the tasks, minAP scores are dramatically increased, e.g., 61.97 vs. 47.41 for the easy 3D detection task, and the maxDiffAP scores are decreased (maxDiffAP scores for AVOD with LEL are reported in Appendix \ref{sec:append_exp}.). Then the robustness is further improved by minimizing our \mssn~loss. As our LEL's structure inherently handles corruption on a single source well, even the \textsc{TrainASN} algorithm can successfully guide the model to be equipped with the desired robustness.

A corruption method with a different style, downsampling, is also tested with our LEL fusion method. Table \ref{table:result_car_lel_full_asn_ssn_downsample_main} shows that the model trained with our \textsc{TrainSSN} algorithm achieves the best robustness among the four algorithms for this complex and realistic perturbation.

\begin{remark}
	A simple \textsc{TrainSSNAlt} achieves fairly robust models in both fusion methods against Gaussian noise, and two reasons may explain this phenomenon. First, all parameters are updated instead of fine-tuning only fusion related parts. Therefore, unlike our analyses on the linear model, the latent representation can be transformed to meet the objective function. In fact, \textsc{TrainSSNAlt} performs poorly when we fine-tune the model with concatenation fusion layers as shown in the supplement. Secondly, the loss function $\cL$ inside our $\Lmssn$ is usually non-convex so that it may be enough to use an indirect approach for small number of sources, $n_s=2$.
\end{remark}

\begin{remark}
	Without applying fancier approaches which could increase computational cost, our LEL showed appealing effectiveness even with simple implementation.
\end{remark}

\section{Conclusion}\label{sec:conclusion}
We study two strategies to improve robustness of fusion models against single source corruption. Motivated by analyses on linear fusion models, a loss function is introduced to balance performance degradation of deep fusion models caused by corruption in different sources. We also demonstrate the importance of a fusion method's structure by proposing a simple ensemble layer achieving such robustness inherently. Our experimental results show that deep fusion models can effectively use complementary and shared information of different input sources by training with our loss and fusion layer to obtain both robustness and high accuracy. We hope our results motivate further work to improve the single source robustness of more complex fusion models with either large number of input sources or adaptive networks. Another interesting direction is to investigate the single source robustness against adversarial attacks in deep fusion models, which can be compared with our analyses in the supplementary material.

%

\begin{small}
\bibliographystyle{plainnat}
\bibliography{kim2019robustfusion}

\begin{thebibliography}{48}
\providecommand{\natexlab}[1]{#1}
\providecommand{\url}[1]{\texttt{#1}}
\expandafter\ifx\csname urlstyle\endcsname\relax
  \providecommand{\doi}[1]{doi: #1}\else
  \providecommand{\doi}{doi: \begingroup \urlstyle{rm}\Url}\fi

\bibitem[Braun et~al.(2016)Braun, Rao, Wang, and Flohr]{braun2016pose}
Markus Braun, Qing Rao, Yikang Wang, and Fabian Flohr.
\newblock Pose-rcnn: Joint object detection and pose estimation using 3d object
  proposals.
\newblock In \emph{IEEE 19th international conference on intelligent
  transportation systems (ITSC)}, pages 1546--1551, 2016.

\bibitem[Chan et~al.(2016)Chan, Jaitly, Le, and Vinyals]{chan2016listen}
William Chan, Navdeep Jaitly, Quoc Le, and Oriol Vinyals.
\newblock Listen, attend and spell: A neural network for large vocabulary
  conversational speech recognition.
\newblock In \emph{IEEE international conference on acoustics, speech and
  signal processing (ICASSP)}, pages 4960--4964, 2016.

\bibitem[Chen et~al.(2017)Chen, Ma, Wan, Li, and Xia]{chen2017multi}
Xiaozhi Chen, Huimin Ma, Ji~Wan, Bo~Li, and Tian Xia.
\newblock Multi-view 3d object detection network for autonomous driving.
\newblock In \emph{IEEE conference on computer vision and pattern recognition
  (CVPR)}, pages 1907--1915, 2017.

\bibitem[Chiu et~al.(2018)Chiu, Sainath, Wu, Prabhavalkar, Nguyen, Chen,
  Kannan, Weiss, Rao, Gonina, et~al.]{chiu2018state}
Chung-Cheng Chiu, Tara~N Sainath, Yonghui Wu, Rohit Prabhavalkar, Patrick
  Nguyen, Zhifeng Chen, Anjuli Kannan, Ron~J Weiss, Kanishka Rao, Ekaterina
  Gonina, et~al.
\newblock State-of-the-art speech recognition with sequence-to-sequence models.
\newblock In \emph{IEEE international conference on acoustics, speech and
  signal processing (ICASSP)}, pages 4774--4778, 2018.

\bibitem[Chorowski et~al.(2015)Chorowski, Bahdanau, Serdyuk, Cho, and
  Bengio]{chorowski2015attention}
Jan~K Chorowski, Dzmitry Bahdanau, Dmitriy Serdyuk, Kyunghyun Cho, and Yoshua
  Bengio.
\newblock Attention-based models for speech recognition.
\newblock In \emph{Advances in neural information processing systems
  (NeurIPS)}, pages 577--585, 2015.

\bibitem[Chung et~al.(2017)Chung, Senior, Vinyals, and Zisserman]{chung2017lip}
Joon~Son Chung, Andrew Senior, Oriol Vinyals, and Andrew Zisserman.
\newblock Lip reading sentences in the wild.
\newblock In \emph{IEEE conference on computer vision and pattern recognition
  (CVPR)}, pages 3444--3453, 2017.

\bibitem[Dai et~al.(2016)Dai, Li, He, and Sun]{dai2016r}
Jifeng Dai, Yi~Li, Kaiming He, and Jian Sun.
\newblock R-fcn: Object detection via region-based fully convolutional
  networks.
\newblock In \emph{Advances in neural information processing systems
  (NeurIPS)}, pages 379--387, 2016.

\bibitem[Du et~al.(2017)Du, Ang, and Rus]{du2017car}
Xinxin Du, Marcelo~H Ang, and Daniela Rus.
\newblock Car detection for autonomous vehicle: Lidar and vision fusion
  approach through deep learning framework.
\newblock In \emph{IEEE/RSJ international conference on intelligent robots and
  systems (IROS)}, pages 749--754, 2017.

\bibitem[Feng et~al.(2019)Feng, Haase-Schuetz, Rosenbaum, Hertlein, Duffhauss,
  Glaeser, Wiesbeck, and Dietmayer]{feng2019deep}
Di~Feng, Christian Haase-Schuetz, Lars Rosenbaum, Heinz Hertlein, Fabian
  Duffhauss, Claudius Glaeser, Werner Wiesbeck, and Klaus Dietmayer.
\newblock Deep multi-modal object detection and semantic segmentation for
  autonomous driving: Datasets, methods, and challenges.
\newblock \emph{arXiv preprint arXiv:1902.07830}, 2019.

\bibitem[Geiger et~al.(2012)Geiger, Lenz, and Urtasun]{geiger2012we}
Andreas Geiger, Philip Lenz, and Raquel Urtasun.
\newblock Are we ready for autonomous driving? the kitti vision benchmark
  suite.
\newblock In \emph{IEEE conference on computer vision and pattern recognition
  (CVPR)}, pages 3354--3361, 2012.

\bibitem[Girshick(2015)]{girshick2015fast}
Ross Girshick.
\newblock Fast r-cnn.
\newblock In \emph{IEEE international conference on computer vision (ICCV)},
  pages 1440--1448, 2015.

\bibitem[Girshick et~al.(2014)Girshick, Donahue, Darrell, and
  Malik]{girshick2014rich}
Ross Girshick, Jeff Donahue, Trevor Darrell, and Jitendra Malik.
\newblock Rich feature hierarchies for accurate object detection and semantic
  segmentation.
\newblock In \emph{IEEE conference on computer vision and pattern recognition
  (CVPR)}, pages 580--587, 2014.

\bibitem[Goodfellow et~al.(2015)Goodfellow, Shlens, and
  Szegedy]{goodfellowexplaining}
Ian~J Goodfellow, Jonathon Shlens, and Christian Szegedy.
\newblock Explaining and harnessing adversarial examples.
\newblock In \emph{International conference on learning representations
  (ICLR)}, 2015.

\bibitem[Graves et~al.(2013)Graves, Mohamed, and Hinton]{graves2013speech}
Alex Graves, Abdel-rahman Mohamed, and Geoffrey Hinton.
\newblock Speech recognition with deep recurrent neural networks.
\newblock In \emph{IEEE international conference on acoustics, speech and
  signal processing (ICASSP)}, pages 6645--6649, 2013.

\bibitem[He et~al.(2016)He, Zhang, Ren, and Sun]{he2016deep}
Kaiming He, Xiangyu Zhang, Shaoqing Ren, and Jian Sun.
\newblock Deep residual learning for image recognition.
\newblock In \emph{IEEE conference on computer vision and pattern recognition
  (CVPR)}, pages 770--778, 2016.

\bibitem[He et~al.(2017)He, Gkioxari, Doll{\'a}r, and Girshick]{he2017mask}
Kaiming He, Georgia Gkioxari, Piotr Doll{\'a}r, and Ross Girshick.
\newblock Mask r-cnn.
\newblock In \emph{IEEE international conference on computer vision (ICCV)},
  pages 2961--2969, 2017.

\bibitem[Hinton et~al.(2012)Hinton, Deng, Yu, Dahl, Mohamed, Jaitly, Senior,
  Vanhoucke, Nguyen, Kingsbury, et~al.]{hinton2012deep}
Geoffrey Hinton, Li~Deng, Dong Yu, George Dahl, Abdel-rahman Mohamed, Navdeep
  Jaitly, Andrew Senior, Vincent Vanhoucke, Patrick Nguyen, Brian Kingsbury,
  et~al.
\newblock Deep neural networks for acoustic modeling in speech recognition.
\newblock \emph{IEEE signal processing magazine}, 29, 2012.

\bibitem[Huang et~al.(2017)Huang, Liu, Van Der~Maaten, and
  Weinberger]{huang2017densely}
Gao Huang, Zhuang Liu, Laurens Van Der~Maaten, and Kilian~Q Weinberger.
\newblock Densely connected convolutional networks.
\newblock In \emph{IEEE conference on computer vision and pattern recognition
  (CVPR)}, pages 4700--4708, 2017.

\bibitem[Huang and Kingsbury(2013)]{huang2013audio}
Jing Huang and Brian Kingsbury.
\newblock Audio-visual deep learning for noise robust speech recognition.
\newblock In \emph{IEEE international conference on acoustics, speech and
  signal processing (ICASSP)}, pages 7596--7599, 2013.

\bibitem[Kim et~al.(2018{\natexlab{a}})Kim, Koh, Kim, Choi, Hwang, and
  Choi]{kim2018robust}
Jaekyum Kim, Junho Koh, Yecheol Kim, Jaehyung Choi, Youngbae Hwang, and Jun~Won
  Choi.
\newblock Robust deep multi-modal learning based on gated information fusion
  network.
\newblock In \emph{Asian conference on computer vision (ACCV)},
  2018{\natexlab{a}}.

\bibitem[Kim and Ghosh(2016)]{kim2016robust}
Taewan Kim and Joydeep Ghosh.
\newblock Robust detection of non-motorized road users using deep learning on
  optical and lidar data.
\newblock In \emph{IEEE 19th international conference on intelligent
  transportation systems (ITSC)}, pages 271--276, 2016.

\bibitem[Kim et~al.(2018{\natexlab{b}})Kim, Motro, Lavieri, Oza, Ghosh, and
  Bhat]{kim2018pedestrian}
Taewan Kim, Michael Motro, Patr{\'\i}cia Lavieri, Saharsh~Samir Oza, Joydeep
  Ghosh, and Chandra Bhat.
\newblock Pedestrian detection with simplified depth prediction.
\newblock In \emph{IEEE 21st international conference on intelligent
  transportation systems (ITSC)}, pages 2712--2717, 2018{\natexlab{b}}.

\bibitem[Kiros et~al.(2014)Kiros, Popuri, Cobzas, and
  Jagersand]{kiros2014stacked}
Ryan Kiros, Karteek Popuri, Dana Cobzas, and Martin Jagersand.
\newblock Stacked multiscale feature learning for domain independent medical
  image segmentation.
\newblock In \emph{International workshop on machine learning in medical
  imaging}, pages 25--32. Springer, 2014.

\bibitem[Krizhevsky et~al.(2012)Krizhevsky, Sutskever, and
  Hinton]{krizhevsky2012imagenet}
Alex Krizhevsky, Ilya Sutskever, and Geoffrey~E Hinton.
\newblock Imagenet classification with deep convolutional neural networks.
\newblock In \emph{Advances in neural information processing systems
  (NeurIPS)}, pages 1097--1105, 2012.

\bibitem[Ku et~al.(2018)Ku, Mozifian, Lee, Harakeh, and Waslander]{ku2018joint}
Jason Ku, Melissa Mozifian, Jungwook Lee, Ali Harakeh, and Steven~L Waslander.
\newblock Joint 3d proposal generation and object detection from view
  aggregation.
\newblock In \emph{IEEE/RSJ international conference on intelligent robots and
  systems (IROS)}, pages 1--8, 2018.

\bibitem[LeCun et~al.(2015)LeCun, Bengio, and Hinton]{lecun2015deep}
Yann LeCun, Yoshua Bengio, and Geoffrey Hinton.
\newblock Deep learning.
\newblock \emph{nature}, 521\penalty0 (7553):\penalty0 436, 2015.

\bibitem[Liang et~al.(2018)Liang, Yang, Wang, and Urtasun]{liang2018deep}
Ming Liang, Bin Yang, Shenlong Wang, and Raquel Urtasun.
\newblock Deep continuous fusion for multi-sensor 3d object detection.
\newblock In \emph{European conference on computer vision (ECCV)}, pages
  641--656, 2018.

\bibitem[Liang et~al.(2019)Liang, Yang, Chen, Hui, and Urtasun]{liang2019multi}
Ming Liang, Bin Yang, Yun Chen, Rui Hui, and Raquel Urtasun.
\newblock Multi-task multi-sensor fusion for 3d object detection.
\newblock In \emph{IEEE conference on computer vision and pattern recognition
  (CVPR)}, 2019.

\bibitem[Liu et~al.(2015)Liu, Liu, Cai, Che, Pujol, Kikinis, Feng, Fulham,
  et~al.]{liu2015multimodal}
Siqi Liu, Sidong Liu, Weidong Cai, Hangyu Che, Sonia Pujol, Ron Kikinis, Dagan
  Feng, Michael~J Fulham, et~al.
\newblock Multimodal neuroimaging feature learning for multiclass diagnosis of
  alzheimer's disease.
\newblock \emph{IEEE transactions on biomedical engineering}, 62\penalty0
  (4):\penalty0 1132--1140, 2015.

\bibitem[Liu et~al.(2016)Liu, Anguelov, Erhan, Szegedy, Reed, Fu, and
  Berg]{liu2016ssd}
Wei Liu, Dragomir Anguelov, Dumitru Erhan, Christian Szegedy, Scott Reed,
  Cheng-Yang Fu, and Alexander~C Berg.
\newblock Ssd: Single shot multibox detector.
\newblock In \emph{European conference on computer vision (ECCV)}, pages
  21--37. Springer, 2016.

\bibitem[Mees et~al.(2016)Mees, Eitel, and Burgard]{mees2016choosing}
Oier Mees, Andreas Eitel, and Wolfram Burgard.
\newblock Choosing smartly: Adaptive multimodal fusion for object detection in
  changing environments.
\newblock In \emph{IEEE/RSJ international conference on intelligent robots and
  systems (IROS)}, pages 151--156, 2016.

\bibitem[Mroueh et~al.(2015)Mroueh, Marcheret, and Goel]{mroueh2015deep}
Youssef Mroueh, Etienne Marcheret, and Vaibhava Goel.
\newblock Deep multimodal learning for audio-visual speech recognition.
\newblock In \emph{IEEE international conference on acoustics, speech and
  signal processing (ICASSP)}, pages 2130--2134, 2015.

\bibitem[Qi et~al.(2018)Qi, Liu, Wu, Su, and Guibas]{qi2018frustum}
Charles~R Qi, Wei Liu, Chenxia Wu, Hao Su, and Leonidas~J Guibas.
\newblock Frustum pointnets for 3d object detection from rgb-d data.
\newblock In \emph{IEEE conference on computer vision and pattern recognition
  (CVPR)}, pages 918--927, 2018.

\bibitem[Ramachandram and Taylor(2017)]{ramachandram2017deep}
Dhanesh Ramachandram and Graham~W Taylor.
\newblock Deep multimodal learning: A survey on recent advances and trends.
\newblock \emph{IEEE signal processing magazine}, 34\penalty0 (6):\penalty0
  96--108, 2017.

\bibitem[Redmon and Farhadi(2017)]{redmon2017yolo9000}
Joseph Redmon and Ali Farhadi.
\newblock Yolo9000: better, faster, stronger.
\newblock In \emph{IEEE conference on computer vision and pattern recognition
  (CVPR)}, pages 7263--7271, 2017.

\bibitem[Redmon et~al.(2016)Redmon, Divvala, Girshick, and
  Farhadi]{redmon2016you}
Joseph Redmon, Santosh Divvala, Ross Girshick, and Ali Farhadi.
\newblock You only look once: Unified, real-time object detection.
\newblock In \emph{IEEE conference on computer vision and pattern recognition
  (CVPR)}, pages 779--788, 2016.

\bibitem[Ren et~al.(2015)Ren, He, Girshick, and Sun]{ren2015faster}
Shaoqing Ren, Kaiming He, Ross Girshick, and Jian Sun.
\newblock Faster r-cnn: Towards real-time object detection with region proposal
  networks.
\newblock In \emph{Advances in neural information processing systems
  (NeurIPS)}, pages 91--99, 2015.

\bibitem[Sainath et~al.(2013)Sainath, Mohamed, Kingsbury, and
  Ramabhadran]{sainath2013deep}
Tara~N Sainath, Abdel-rahman Mohamed, Brian Kingsbury, and Bhuvana Ramabhadran.
\newblock Deep convolutional neural networks for lvcsr.
\newblock In \emph{IEEE international conference on acoustics, speech and
  signal processing (ICASSP)}, pages 8614--8618, 2013.

\bibitem[Simonovsky et~al.(2016)Simonovsky, Guti{\'e}rrez-Becker, Mateus,
  Navab, and Komodakis]{simonovsky2016deep}
Martin Simonovsky, Benjam{\'\i}n Guti{\'e}rrez-Becker, Diana Mateus, Nassir
  Navab, and Nikos Komodakis.
\newblock A deep metric for multimodal registration.
\newblock In \emph{International conference on medical image computing and
  computer-assisted intervention}, pages 10--18. Springer, 2016.

\bibitem[Simonyan and Zisserman(2015)]{simonyan2014very}
Karen Simonyan and Andrew Zisserman.
\newblock Very deep convolutional networks for large-scale image recognition.
\newblock In \emph{International conference on learning representations
  (ICLR)}, 2015.

\bibitem[Sui et~al.(2015)Sui, Bennamoun, and Togneri]{sui2015listening}
Chao Sui, Mohammed Bennamoun, and Roberto Togneri.
\newblock Listening with your eyes: Towards a practical visual speech
  recognition system using deep boltzmann machines.
\newblock In \emph{IEEE international conference on computer vision (ICCV)},
  pages 154--162, 2015.

\bibitem[Szegedy et~al.(2015)Szegedy, Liu, Jia, Sermanet, Reed, Anguelov,
  Erhan, Vanhoucke, and Rabinovich]{szegedy2015going}
Christian Szegedy, Wei Liu, Yangqing Jia, Pierre Sermanet, Scott Reed, Dragomir
  Anguelov, Dumitru Erhan, Vincent Vanhoucke, and Andrew Rabinovich.
\newblock Going deeper with convolutions.
\newblock In \emph{IEEE conference on computer vision and pattern recognition
  (CVPR)}, pages 1--9, 2015.

\bibitem[Tsipras et~al.(2019)Tsipras, Santurkar, Engstrom, Turner, and
  Madry]{tsipras2018robustness}
Dimitris Tsipras, Shibani Santurkar, Logan Engstrom, Alexander Turner, and
  Aleksander Madry.
\newblock Robustness may be at odds with accuracy.
\newblock In \emph{International conference on learning representations
  (ICLR)}, 2019.

\bibitem[Tumer and Ghosh(1996{\natexlab{a}})]{tumer1996analysis}
Kagan Tumer and Joydeep Ghosh.
\newblock Analysis of decision boundaries in linearly combined neural
  classifiers.
\newblock \emph{Pattern Recognition}, 29\penalty0 (2):\penalty0 341--348,
  1996{\natexlab{a}}.

\bibitem[Tumer and Ghosh(1996{\natexlab{b}})]{tumer1996error}
Kagan Tumer and Joydeep Ghosh.
\newblock Error correlation and error reduction in ensemble classifiers.
\newblock \emph{Connection science}, 8\penalty0 (3-4):\penalty0 385--404,
  1996{\natexlab{b}}.

\bibitem[Valada et~al.(2017)Valada, Vertens, Dhall, and
  Burgard]{valada2017adapnet}
Abhinav Valada, Johan Vertens, Ankit Dhall, and Wolfram Burgard.
\newblock Adapnet: Adaptive semantic segmentation in adverse environmental
  conditions.
\newblock In \emph{IEEE international conference on robotics and automation
  (ICRA)}, pages 4644--4651, 2017.

\bibitem[Wang et~al.(2018)Wang, Zhan, and Tomizuka]{wang2018fusing}
Zining Wang, Wei Zhan, and Masayoshi Tomizuka.
\newblock Fusing bird’s eye view lidar point cloud and front view camera
  image for 3d object detection.
\newblock In \emph{IEEE intelligent vehicles symposium (IV)}, pages 1--6, 2018.

\bibitem[Wu et~al.(2013)Wu, Hoi, Xia, Zhao, Wang, and Miao]{wu2013online}
Pengcheng Wu, Steven~CH Hoi, Hao Xia, Peilin Zhao, Dayong Wang, and Chunyan
  Miao.
\newblock Online multimodal deep similarity learning with application to image
  retrieval.
\newblock In \emph{21st ACM international conference on multimedia}, pages
  153--162. ACM, 2013.

\end{thebibliography}
\end{small}
\newpage
\clearpage
\appendix
\section{Proofs and supplementary Analyses}\label{sec:append_proof}

\subsection{Proofs and analyses for Section \ref{subsec:siloss}}

\begin{proof}
The original $\Lmssn$ loss minimization problem with an additional constraint of preserving loss under clean data can be transformed to the problem stated in (\ref{eq:opt_rand}) due to the flexibility of $g_1$ and $g_2$ under the constraint $g_1+g_2=\beta_3$:
\begin{equation*}
\min_{g_1,g_2}\max\left\{\cL\left(y,\fdir(x_1+\delta_1,x_2)\right) ,\cL\left(y,\fdir(x_1,x_2+\delta_2)\right)\right\}\hspace{1em}s.t.~~g_1+g_2=\beta_3
\end{equation*}
Under the expected squared loss with $\fdir$ function, the loss can be evaluated,
\begin{align*}
\cL\left(y,\fdir(x_1+\delta_1,x_2)\right) &= \E\left[\left( y-(\beta_1^T (z_1 + \epsilon_1) + g_1^T (z_3 + \epsilon_3) + \beta_2^T z_2 + g_2^T z_3) \right)^2\right]\\
&=\E\left[\left( \beta_1^T \epsilon_1 + g_1^T \epsilon_3\right)^2\right] \hspace{4em}\left(\because y=\sum_{i=1}^3 {\beta_i} ^T z_i\right)\\
&= \sigma^2 (||\beta_1||_2^2 + ||g_1||_2^2) \hspace{4em}(\because \text{Statistical assumption on }\epsilon_i.)
\end{align*}
Hence the equivalent problem (\ref{eq:opt_rand_reduced}) is achieved.
\begin{equation}\label{eq:opt_rand_reduced}
\sigma^2\min_{g_1,g_2}\max\left\{||\beta_1||_2^2+||g_1||_2^2,||\beta_2||_2^2+||g_2||_2^2 \right\}\hspace{1em}s.t.~~g_1+g_2=\beta_3
\end{equation}
For simple notation, substitute variables as $g=g_1,v=\beta_3,c_1=||\beta_1||_2^2,c_2=||\beta_2||_2^2$, and solve the following convex optimization problem.
\begin{equation*}
\min_g \max\{||g||_2^2+c_1,||g-v||_2^2+c_2 \}
\end{equation*}
This problem can be solved by introducing a variable $\gamma$ for the upper bound of the inner maximum value:
\begin{equation*}
\min_{g,\gamma}\gamma\hspace{1em}s.t.~~c_1+||g||_2^2-\gamma\leq0,~c_2+||g-v||_2^2-\gamma\leq 0
\end{equation*}

KKT condition gives:
\begin{align*}
\text{(Primal feasibility)}\hspace{2em}&c_1+||g||_2^2-\gamma\leq 0,~c_2+||g-v||_2^2-\gamma\leq 0\\
\text{(Dual feasibility)}\hspace{2em}&\lambda_1\geq 0,~\lambda_2\geq 0\\
\text{(Complementary slackness)}\hspace{2em}&\lambda_1(c_1+||g||_2^2-\gamma)= 0,~\lambda_2(c_2+||g-v||_2^2-\gamma)= 0\\
\text{(Stationary)}\hspace{2em}&\lambda_1+\lambda_2=1,~g=\frac{\lambda_2}{\lambda_1+\lambda_2}v
\end{align*}
Considering $\lambda_1+\lambda_2=1$ and $\lambda_1,\lambda_2\geq0$, we first need to analyze the case $\lambda_1=0$. This gives $g=v$ and the complementary slackness condition to find $\gamma=c_2+||g-v||_2^2=c_2$. $\lambda_2=0$ can be analyzed with similar steps. If both $\lambda_1$ and $\lambda_2$ are positive, the complementary slackness condition gives $\gamma=c_1 +||g||_2^2 = c_2 + ||g-v||_2^2$, which ensures the balance of the original problem's maximum value $\max\{c_1 +||g||_2^2,c_2 + ||g-v||_2^2\}$. This case gives $\gamma=\frac{c_1+c_2}{2}+\frac{||v||_2^2}{4}+\frac{(c_2-c_1)^2}{4||v||_2^2}$ with $g=\frac{1}{2}\left(1+\frac{c_2-c_1}{||v||_2^2} \right)v$. Therefore, we can have the result (\ref{eq:sol_linear_rand}) which provides the fusion model robust against single source corruptions from random noise.
\end{proof}

\paragraph{Comparison to the model not considering \mssn~loss}If random noise are added to $x_1$ and $x_2$ simultaneously, the objective of the problem becomes $\min_{g_1,g_2} \E[(y-\fdir(x_1+\delta_1,x_2+\delta_2))^2]$ instead of considering the $\mssn$~loss. This is equivalent to minimizing $\sigma^2(||\beta_1||_2^2 + ||\beta_2||_2^2+||g_1||_2^2+||g_2||_2^2 )$ subject to $g_1+g_2=\beta_3$, and the solution can be directly found as it is a simple convex problem, which is $g_1=g_2=\frac{\beta_3}{2}$. If we denote this model as $\fdir'$, then \mssn~ loss is:
\begin{equation*}
\Lmssn(\fdir',\epsilon)=\Lmssn'=\sigma^2\max\left\{||\beta_1||_2^2+\frac{1}{4}||\beta_3||_2^2,||\beta_2||_2^2+\frac{1}{4}||\beta_3||_2^2 \right\}
\end{equation*}
Now, let's compute the difference $\Lmssn'-\Lmssn^*$.
\begin{proof}
As both term includes $\sigma^2$, let's assume $\sigma^2=1$ for ease of notation. Among the three cases in (\ref{eq:sol_linear_rand}), consider the first case $||\beta_1||_2^2+||\beta_3||_2^2\leq ||\beta_2||_2^2$.
\begin{align*}
\Lmssn'-\Lmssn^*=||\beta_2||_2^2+\frac{1}{4}||\beta_3||_2^2-||\beta_2||_2^2 = \frac{1}{4}||\beta_3||_2^2 \hspace{1em}(\because ||\beta_2||_2^2 \geq ||\beta_1||_2^2)
\end{align*}
The second case can be shown similarly. Now assume that $\left|\frac{||\beta_2||_2^2-||\beta_1||_2^2}{||\beta_3||_2^2}\right| < 1$ holds, and let $||\beta_2||_2^2 \geq ||\beta_1||_2^2$ without loss of generality. Then we can show that,
\begin{align*}
\Lmssn' - \Lmssn^* &= ||\beta_2||_2^2+\frac{1}{4}||\beta_3||_2^2-\left(\frac{||\beta_1||_2^2+||\beta_2||_2^2}{2}+\frac{||\beta_3||_2^2}{4}+\frac{(||\beta_2||_2^2-||\beta_1||_2^2)^2}{4||\beta_3||_2^2}\right)\\
&=\frac{1}{2}(||\beta_2||_2^2-||\beta_1||_2^2)\left(1-\frac{||\beta_2||_2^2-||\beta_1||_2^2}{2||\beta_3||_2^2} \right)\\
&\geq \frac{1}{4}(||\beta_2||_2^2-||\beta_1||_2^2)\hspace{1em}\left(\because ||\beta_2||_2^2 \geq ||\beta_1||_2^2\text{ and }\left|\frac{||\beta_2||_2^2-||\beta_1||_2^2}{||\beta_3||_2^2}\right| < 1 \right)
\end{align*}
\end{proof}
Therefore we can conclude that simply optimizing under noise added to all the input sources at the same time cannot do better than minimizing \mssn~loss with some nonnegative gap in our linear fusion model.

\subsection{Single Source Robustness against Adversarial attacks}\label{subsec:adv_attack}
Another important type of perturbation is an adversarial attack. Different from the previously studied random noise, perturbation to the input sources is also optimized to maximize the loss to consider the worst case. Adversarial version of the \mssn~loss is defined as follows:

\begin{defn}\label{dfn:adv_maxssn_loss}
	For multiple sources $x_1,\cdots,x_{n_s}$ and a target variable $y$, denote a predefined loss function by $\cL$. If each input source $x_i$ is maximally perturbed with some additive noise $\eta_i\in \cS_i$ for $i\in[n_s]$, \mssnadv~loss for a model $f$ is defined as follows:
	\begin{align*}
	\Lmssnadv(f,\eta)\triangleq \max_{i} \left\{ \max_{\eta_i\in \cS_i} \cL\left( y,f(x_i+\eta_i,x_{-i}) \right)\right\}_{i=1}^{n_s}
	\end{align*}
\end{defn}

As a simple model analysis, let's consider a binary classification problem using the logistic regression. Again, two input sources $x_1=[z_1;z_3]$ and $x_2=[z_2;z_3]$ have a common feature vector $z_3$ as in the linear fusion data model. A binary classifier $\sgn(f(x_1,x_2))$ is trained to predict label $y\in\{-1,1\}$, where $f(x_1,x_2)=(w_1^Tz_1+g_1^Tz_3)+(w_2^Tz_2+g_2^Tz_3)$ and the training loss is $\E_{x,y}\left[ \ell\left( y\cdot f(x_1,x_2) \right) \right]$ with the logistic function $\ell(x)=\log(1+\exp(-x))$. Here, we apply one of the most popular attacks, fast gradient sign (FGS) method, which was also motivated by linear models without a fusion framework \citep{goodfellowexplaining}. The adversarial attack $\eta_i$ per each source $x_i$ under $\ell_\infty$ norm constraint $||\eta_i||_\infty\leq \varepsilon$ can be similarly derived as follows:
\begin{equation}\label{eq:adv_attack_eta}
\eta_1 = [-\varepsilon y\cdot\sgn(w_1);-\varepsilon y\cdot\sgn(g_1)],~\eta_2=[-\varepsilon y\cdot\sgn(w_2);-\varepsilon y\cdot\sgn(g_2)]
\end{equation}
As a substitute for the linear fusion data model, let's assume the true classes are generated by the hidden relationship $y=\sgn\left(\sum_{i=1}^{3}\beta_i^T z_i\right)$. Then the optimal fusion binary classifier becomes $\sgn(\fdir(x_1,x_2))$. Similar to the previous section, suppose an objective is to find a model with robustness against single source adversarial attacks, while preserving the performance on clean data. Then the overall optimization problem can be reduced to the following one:
\begin{equation}\label{eq:opt_adv}
\min_{g_1,g_2}\max\left\{\cL\left(y,\fdir(x_1+\eta_1,x_2)\right) ,\cL\left(y,\fdir(x_1,x_2+\eta_2)\right)\right\}\hspace{1em}s.t.~~g_1+g_2=\beta_3
\end{equation}
As $\ell$ is a decreasing function, optimal $g_1$ and $g_2$ of the original problem are equivalent to the minimizer of the following one:
\begin{equation}\label{eq:opt_adv_reduced}
\varepsilon\min_{g_1,g_2}\max\left\{ ||w_1||_1+||g_1||_1, ||w_2||_1+||g_2||_1 \right\}\hspace{1em}s.t.~~g_1+g_2=\beta_3
\end{equation}
By solving this convex optimization problem, we can achieve solution $\Lmssnadv^*$ and optimizers $g_1^*,g_2^*$. Also, we can find $\Lmssnadv'$, a $\Lmssnadv$ value evaluated using the optimal model for minimizing the adversarial attacks added to all the sources at once. Interestingly, we can show that $\Lmssnadv'\geq\Lmssnadv^*$ if $\frac{||\beta_2||_1 - ||\beta_1||_1}{||\beta_3||_1}>1$, but $\Lmssnadv'=\Lmssnadv^*$ otherwise. In other words, if inherent influence of $z_1$ and $z_2$ are well balanced compared to the common feature $z_3$ in the sense of $\ell_1$ norm, adversarial attacks only applied to a single source can be equivalently defended by just using a traditional adversarial training strategy to learn a model robust against attacks added to all the sources at once.

\begin{proof}
The original minimizing $\Lmssnadv$ loss minimization problem with an additional constraint of preserving loss under clean data can be transformed to the problem stated in (\ref{eq:opt_adv}) due to the flexibility of $g_1$ and $g_2$:
\begin{equation*}
\min_{g_1,g_2}\max\left\{\cL\left(y,\fdir(x_1+\eta_1,x_2)\right) ,\cL\left(y,\fdir(x_1,x_2+\eta_2)\right)\right\}\hspace{1em}s.t.~~g_1+g_2=\beta_3
\end{equation*}
As $\eta_i$'s are assumed to be made with FGS method, adversarial attacks under $\ell_\infty$ norm constraints are as follows:
\begin{equation*}
\eta_1 = [-\varepsilon y\cdot\sgn(w_1);-\varepsilon y\cdot\sgn(g_1)],~\eta_2=[-\varepsilon y\cdot\sgn(w_2);-\varepsilon y\cdot\sgn(g_2)]
\end{equation*}
Therefore, minimizing $\Lmssnadv(\fdir,\eta)$ over $g_1,g_2$ becomes: 
\begin{align*}
\min_{g_1,g_2}\max \{&\E\left[\ell\left( y\cdot \fdir(x_1,x_2)-\varepsilon(||w_1||_1+||g_1||_1) \right)\right],\\
&\E\left[\ell\left( y\cdot \fdir(x_1,x_2)-\varepsilon(||w_2||_1+||g_2||_1) \right)\right] \}\hspace{1em}s.t.~~g_1+g_2=\beta_3
\end{align*}

We can solve the following problem to find minimizers $g_1^*$ and $g_2^*$.
\begin{equation*}
\min_{g_1,g_2}\max\left\{ ||w_1||_1+||g_1||_1, ||w_2||_1+||g_2||_1 \right\}\hspace{1em}s.t.~~g_1+g_2=\beta_3
\end{equation*}
Similar to the random noise case, substitute variables as $g=g_1,v=\beta_3,c_1=||\beta_1||_1,c_2=||\beta_2||_2$, and solve the following convex optimization problem:
\begin{equation*}
\min_g \max\{||g||_1+c_1,||g-v||_1+c_2 \}
\end{equation*}
which can be solved by introducing $\gamma$,
\begin{equation*}
\min_{g,\gamma}\gamma\hspace{1em}s.t.~~c_1+||g||_1-\gamma\leq0,~c_2+||g-v||_1-\gamma\leq 0
\end{equation*}

KKT condition gives:
\begin{align*}
\text{(Primal feasibility)}\hspace{2em}&c_1+||g||_1-\gamma\leq 0,~c_2+||g-v||_1-\gamma\leq 0\\
\text{(Dual feasibility)}\hspace{2em}&\lambda_1\geq 0,~\lambda_2\geq 0\\
\text{(Complementary slackness)}\hspace{2em}&\lambda_1(c_1+||g||_1-\gamma)= 0,~\lambda_2(c_2+||g-v||_1-\gamma)= 0\\
\text{(Stationary)}\hspace{2em}&\lambda_1+\lambda_2=1,~0 \in \lambda_1\partial ||g||_1 +\lambda_2\partial||g-v||_1
\end{align*}
If $\lambda_1=0$ or $\lambda_2=0$, these cases handle when the inherent imbalance of three components $z_1,z_2$ and $z_3$. Consider $\lambda_2=0$, which gives $||g||_1+c_1-\gamma=0$ from the complementary slackness condition. And the stationary condition becomes $0\in \partial ||g||_1$. As a subgradient of $||g||_1$ can be zero if and only if $g(i)=0$ for any $i^{th}$ component, the solution is $g=0$ with $\gamma=c_1$ and the necessary condition is $||v||_1+c_2\leq c_1$. Similar solution can be found for $\lambda_1=0$ case as $g=v,\gamma=c_2$ if $||v||_1+c_1\leq c_2$. Therefore, we can have $\gamma^*=\min\max\left\{ ||w_1||_1+||g_1||_1, ||w_2||_1+||\beta_3-g_1||_1 \right\}$ and corresponding parameters as:
\begin{equation*}
\left(\gamma^*,g_1^*,g_2^*\right)=\begin{cases}
\left(||\beta_2||_1,\beta_3,0 \right) & \text{if } ||\beta_1||_1+||\beta_3||_1\leq ||\beta_2||_1\\
\left(||\beta_1||_1,0,\beta_3 \right) & \text{if } ||\beta_2||_1+||\beta_3||_1\leq ||\beta_1||_1
\end{cases}
\end{equation*}

Now let's consider $\lambda_1\neq0,\lambda_2\neq0$. Denote $q \in \lambda_1\partial ||g||_1 +\lambda_2\partial||g-v||_1$ as the element of subdifferential of the Lagrangian. We need to find cases for $q(i)=0$ to hold. 

(i) If $v(i)=0$, then $\sgn(g(i))=\sgn(g(i)-v(i))$ holds. Therefore, if $g(i)\neq=0$, a subgradient becomes $q(i)=\lambda_1 \sgn(g(i))+\lambda_2\sgn(g(i))=\sgn(g(i))$ which cannot be zero. $\Rightarrow$ $g(i)=v(i)=0$.

(ii) If $v(i)\neq 0$, we need to consider three different sub cases. First, if $g(i)\neq 0$ and $g(i)\neq v(i)$, then $q(i)=\lambda_1(\sgn(g(i))-\sgn(g(i)-v(i)))+\sgn(g(i)-v(i))$. For $q(i)=0$ to hold, $\sgn(g(i))=-\sgn(g(i)-v(i))$ must be true with $\lambda_1=\frac{1}{2}$. This gives a solution $g(i)=\alpha_i v(i)$ with $\forall\alpha_i\in(0,1)$.

Secondly, if $g(i)=0$ but $g(i)\neq v(i)$, then the subgradient is $q(i)=\lambda_1 \alpha_i + (1-\lambda_1)\sgn(-v(i))$ for any $\alpha_i\in[-1,1]$. Therefore, if $\alpha_i=\frac{1-\lambda_1}{\lambda_1}\sgn(v(i))$ with some $\lambda_1\in[\frac{1}{2},1)$, the stationary condition holds.

Finally, if $g(i)\neq 0$ and $g(i)=v(i)$, then $q(i)=\lambda_1 \sgn(g(i))+(1-\lambda_1)\alpha_i$ for any $\alpha_i \in [-1,1]$. Therefore, if $\alpha_i=\frac{\lambda_1}{1-\lambda_1}\sgn(v(i))$ with $\lambda_1\in(0,\frac{1}{2}]$, $q(i)=0$ holds for the stationary condition.

All the above cases in (i) and (ii) can be restated as a combined solution $g(i)=\alpha_i v(i)$, $\forall \alpha_i \in [0,1]$. It is easy to show that $|g(i)|+|g(i)-v(i)|=|v(i)|$ holds for any $i$. Also, $\lambda_1\neq0,\lambda_2\neq0$ with the complementary slackness condition gives a new constraint $\gamma=||g||_1+c_1 = ||g-v||_1+c_2$. Hence, we can calculate $\gamma$ by averaging the two equivalent values:
\begin{equation*}
\gamma=\frac{1}{2}(c_1+c_2+||g||_1+||g-v||_1)=\frac{1}{2}(c_1+c_2+||v||_1)
\end{equation*}
Therefore, $(\gamma^*,g_1^*,g_2^*)=\left(\frac{1}{2}(||\beta_1||_1+||\beta_2||_1+||\beta_3||_1), \alpha\odot\beta_3 , \beta_3-\alpha\odot\beta_3 \right)$, where $\odot$ is an element-wise product and each element of $\alpha$ can have any value in $[0,1]$, i.e. $\alpha(i)\in[0,1]$.

Now, let's consider a model robust against adversarial attacks added to both sources $x_1$ and $x_2$ at the same time. This becomes a problem of minimizing $||\beta_1||_1+||\beta_2||_1+||g_1||_1+||\beta_3-g_1||_1$. And the optimal solution can be achieved by $(g_1',g_2')=(\alpha\odot\beta_3 , \beta_3-\alpha\odot\beta_3)$ for any alpha satisfying $\alpha(i)\in[0,1]$. Therefore, we can conclude that our $\Lmssnadv$ loss is necessary to give a binary classifier more robust against single source adversarial attacks, i.e. $\Lmssnadv^* \leq \Lmssnadv'$, if $\frac{||\beta_2||_1 - ||\beta_1||_1}{||\beta_3||_1}>1$ holds. Surprisingly, if $\frac{||\beta_2||_1 - ||\beta_1||_1}{||\beta_3||_1}\leq1$ holds to have balanced influence from inherent components from the different source of inputs, 
$\Lmssnadv^* = \Lmssnadv'$. In other words, if different input sources contributes to the target variable with certain balance, a traditional way of generating adversarial samples by considering all the sources at once can train a model robust against single source attacks as well.
\end{proof}

\section{Additional Experimental Results}\label{sec:append_exp}
\paragraph{Evaluation on ASN data}
Although our main focus is corruption on a single source, it is possible for a model to encounter a case where all the sources are corrupted. If the level of corruption is severe, then extracting any meaningful information from the input sources is impossible, e.g. occlusion on every sensors. However, we hope our model to be robust against reasonably corrupted input sources even if our training objective leans toward the single source robustness. Therefore, we also report the model's performance against data corrupted with ASN. In most cases, the AVOD learned with \textsc{TrainASN} method achieves the best robustness against ASN, which is designed to do so. However, a model using element-wise mean fusion layers trained with \textsc{TrainASN} shows lower robustness scores compared to the SSN oriented approaches. We believe that this phenomenon is caused by corrupted feature extraction combined with the structural limitation of the mean fusion layer.

\paragraph{Fine-tuning} We also consider another algorithmic framework using \textit{fine-tuing}. The algorithm starts with a normal training on clean data for $\numclean$ iterations, which may include some general data augmentation methods like random cropping, and flipping. Then $\numtune$ steps of fine-tuning is run to update only a subset of the model's parameters, $\thetafusion\subset f$, so that any essential parts for extracting features from normal data are not affected. Convolutional layers extracting features from different sources before the fusion stages are fixed, and other layers for fusing the features and making predictions are updated in the fine-tuning stage. The experimental results using this method are provided in Table \ref{table:result_car_simple_rand_tune_asn_ssn} and \ref{table:result_car_lel_rand_tune_asn_ssn} for the Gaussian noise case. Overall performance of the fusion model trained from the scratch is better than using fine-tuning. This shows the importance of feature extraction parts in deep learning models.



\begin{table}[th]
\caption{Car detection (3D/BEV) performance of AVOD with \textit{element-wise mean} fusion layers against \textit{Gaussian} \ssn~and \asn~on the KITTI validation set.}
\label{table:result_car_simple_full_asn_ssn}
\begin{center}
\begin{scriptsize}
\begin{tabular}{lcccccc}
\toprule
	(Data) Train Algo. & Easy & Moderate & Hard & Easy & Moderate & Hard\\
\midrule
	(Clean Data) & \multicolumn{3}{c}{$\APTD (\%)$} & \multicolumn{3}{c}{$\APBEV (\%)$}\\ 
	\cmidrule(r){2-4} \cmidrule(r){5-7}
	AVOD \citep{ku2018joint} & 76.41 & \bf{72.74} & \bf{66.86} & 89.33 & 86.49 & \bf{79.44}\\
	+\textsc{TrainASN} & 75.96 & 66.68 & 65.97 & 88.63 & 79.45 & 78.79\\
	+\textsc{TrainSSN} & 76.28 & 67.10 & 66.51 & 88.86 & 79.60 & 79.11\\
	+\textsc{TrainSSNAlt} & \textbf{77.46} & 67.61 & 66.06 & \bf{89.68} & \bf{86.71} & 79.41\\
\midrule
	(Gaussian ASN) & \multicolumn{3}{c}{$\APTD (\%)$} & \multicolumn{3}{c}{$\APBEV (\%)$}\\
	\cmidrule(r){2-4} \cmidrule(r){5-7}
	AVOD \citep{ku2018joint} & $28.08 {\pm} 0.91$ & $26.35 {\pm} 2.18$ & $21.81 {\pm} 0.63$ & $42.01 {\pm} 0.23$ & $33.68 {\pm} 0.17$ & $33.60 {\pm} 0.13$ \\
	+\textsc{TrainASN} & $61.26 {\pm} 0.45$ & $47.71 {\pm} 0.24$ & $45.60 {\pm} 0.19$ & $87.40 {\pm} 0.07$ & $72.07 {\pm} 2.89$ & $70.13 {\pm} 0.05$\\
	+\textsc{TrainSSN} & $69.33 {\pm} 0.43$ & $55.41 {\pm} 0.21$ & $\mathbf{52.90 {\pm} 2.12}$ & $\mathbf{88.39 {\pm} 0.13}$ & $\mathbf{78.37 {\pm} 0.10}$ & $\mathbf{70.75 {\pm} 0.05}$\\
	+\textsc{TrainSSNAlt} & $\mathbf{71.63 {\pm} 0.04}$ & $\mathbf{56.24 {\pm} 0.16}$ & $49.14 {\pm} 0.10$ & $87.95 {\pm} 0.08$ & $77.88 {\pm} 0.17$ & $69.96 {\pm} 0.08$ \\
\midrule
	(Gaussian SSN) & \multicolumn{3}{c}{$\min\APTD (\%)$} & \multicolumn{3}{c}{$\min\APBEV (\%)$}\\
	\cmidrule(r){2-4} \cmidrule(r){5-7}
	AVOD \citep{ku2018joint} & $47.41 {\pm} 0.28$ & $41.84 {\pm} 0.17$ & $36.47 {\pm} 0.16$ & $65.63 {\pm} 0.28$ & $58.02 {\pm} 0.23$ & $50.43 {\pm} 0.14$\\
	+\textsc{TrainASN} & $61.53 {\pm} 0.57$ & $52.72 {\pm} 0.08$ & $47.25 {\pm} 0.13$ & $87.71 {\pm} 0.14 $ & $78.37 {\pm} 0.06$ & $77.85 {\pm} 0.08$ \\
	+\textsc{TrainSSN} & $71.65 {\pm} 0.31$ & $\mathbf{62.14 {\pm} 0.08}$ & $\mathbf{56.78 {\pm} 0.12}$ & $88.21 {\pm} 0.08$ & $78.90 {\pm} 0.09$ & $\mathbf{77.92 {\pm} 0.11}$\\
	+\textsc{TrainSSNAlt} & $\mathbf{71.66 {\pm} 0.48}$ & $57.61 {\pm} 0.12$ & $55.90 {\pm} 0.11$
	& $\mathbf{89.42 {\pm} 0.04}$ & $\mathbf{79.56 {\pm} 0.06}$ & $\mathbf{77.92 {\pm} 0.05}$\\
	\midrule
	(Gaussian SSN) & \multicolumn{3}{c}{$\max \text{Diff}\APTD (\%)$} & \multicolumn{3}{c}{$\max \text{Diff}\APBEV (\%)$}\\
	AVOD \citep{ku2018joint} & $26.70 {\pm} 0.52$ & $22.42 {\pm} 0.29$ & $20.92 {\pm} 0.25$ & $22.27 {\pm} 0.41$ & $20.76 {\pm} 0.33$ & $20.09 {\pm} 0.20$\\
	+\textsc{TrainASN} & $14.48 {\pm} 0.82$ & $12.72 {\pm} 0.33$ & $11.18 {\pm} 0.27$ & $0.88 {\pm} 0.22$ & $0.48 {\pm} 0.13$ & $0.28 {\pm} 0.12$ \\
	+\textsc{TrainSSN} & $\mathbf{3.71 {\pm} 0.46}$ & $\mathbf{3.42 {\pm} 0.25}$ & $7.50 {\pm} 0.25$ & $0.36 {\pm} 0.17$ & $\mathbf{0.04 {\pm} 0.15}$ & $0.71 {\pm} 0.17$\\
	+\textsc{TrainSSNAlt} & $5.55 {\pm} 0.81$ & $8.73 {\pm} 0.32$ & $\mathbf{2.91 {\pm} 0.22}$ & $\mathbf{0.09 {\pm} 0.14}$ & $0.13 {\pm} 0.11$ & $\mathbf{0.18 {\pm} 0.11}$\\
	\bottomrule
\end{tabular}
\end{scriptsize}
\end{center}
\end{table}

\begin{table}[th]
	\caption{Car detection (3D/BEV) performance of AVOD with \textit{element-wise mean} fusion layers (trained with fine-tuning) against \textit{Gaussian} \ssn~and \asn~on the KITTI validation set.}
	\label{table:result_car_simple_rand_tune_asn_ssn}
	\begin{center}
		\begin{scriptsize}
			\begin{tabular}{lcccccc}
				\toprule
				(Data) Train Algo. & Easy & Moderate & Hard & Easy & Moderate & Hard\\
				\midrule
				(Clean Data) & \multicolumn{3}{c}{$\APTD (\%)$} & \multicolumn{3}{c}{$\APBEV (\%)$}\\ 
				\cmidrule(r){2-4} \cmidrule(r){5-7}
				AVOD \citep{ku2018joint} & \bf{76.41} & \bf{72.74} & \bf{66.86} & \bf{89.33} & \bf{86.49} & \bf{79.44}\\
				+\textsc{TrainASN} & 62.55 & 55.81 & 55.34 & 79.08 & 69.90 & 69.83\\
				+\textsc{TrainSSN} & 73.50 & 65.66 & 64.74 & 88.27 & 85.65 & 78.98\\
				+\textsc{TrainSSNAlt} & 75.76 & 71.99 & 66.31 & 88.76 & 85.73 & 79.14\\
				\midrule
				(Gaussian ASN) & \multicolumn{3}{c}{$\APTD (\%)$} & \multicolumn{3}{c}{$\APBEV (\%)$}\\
				\cmidrule(r){2-4} \cmidrule(r){5-7}
				AVOD \citep{ku2018joint} & $28.08 {\pm} 0.91$ & $26.35 {\pm} 2.18$ & $21.81 {\pm} 0.63$ & $42.01 {\pm} 0.23$ & $33.68 {\pm} 0.17$ & $33.60 {\pm} 0.13$ \\
				+\textsc{TrainASN} & $\mathbf{68.58 {\pm} 1.93}$ & $\mathbf{54.76 {\pm} 0.30}$ & $\mathbf{48.00 {\pm} 0.29}$ & $\mathbf{83.15 {\pm} 3.01}$ & $\mathbf{76.10 {\pm} 0.069}$ & $\mathbf{68.49 {\pm} 0.08}$\\
				+\textsc{TrainSSN} & $60.73 {\pm} 0.32$ & $45.52 {\pm} 0.19$ & $44.42 {\pm} 0.11$ & $78.24 {\pm} 0.10$ & $68.41 {\pm} 0.10$ & $60.45 {\pm} 0.07$\\
				+\textsc{TrainSSNAlt} & $53.25 {\pm} 0.27$ & $44.96 {\pm} 0.14$ & $38.64 {\pm} 0.10$ & $68.69 {\pm} 0.18$ & $59.41 {\pm} 0.14$ & $51.37 {\pm} 0.07$ \\
				\midrule
				(Gaussian SSN) & \multicolumn{3}{c}{$\min\APTD (\%)$} & \multicolumn{3}{c}{$\min\APBEV (\%)$}\\
				\cmidrule(r){2-4} \cmidrule(r){5-7}
				AVOD \citep{ku2018joint} & $47.41 {\pm} 0.28$ & $41.84 {\pm} 0.17$ & $36.47 {\pm} 0.16$ & $65.63 {\pm} 0.28$ & $58.02 {\pm} 0.23$ & $50.43 {\pm} 0.14$\\
				+\textsc{TrainASN} & $52.72 {\pm} 0.34$ & $45.66 {\pm} 0.24$ & $39.29 {\pm} 0.22$ & $69.33 {\pm} 0.21$ & $60.19 {\pm} 0.15$ & $59.66 {\pm} 0.15$ \\
				+\textsc{TrainSSN} & $62.46 {\pm} 0.48$ & $53.85 {\pm} 0.22$ & $47.62 {\pm} 0.14$ & $77.77 {\pm} 0.16$ & $68.71 {\pm} 0.09$ & $67.89 {\pm} 0.09$\\
				+\textsc{TrainSSNAlt} & $\mathbf{70.09 {\pm} 0.46}$ & $\mathbf{56.20 {\pm} 0.21}$ & $\mathbf{54.46 {\pm} 0.13}$
				& $\mathbf{84.46 {\pm} 2.66}$ & $\mathbf{76.32 {\pm} 0.06}$ & $\mathbf{68.74 {\pm} 0.08}$\\
				\bottomrule
			\end{tabular}
		\end{scriptsize}
	\end{center}
\end{table}

\begin{table}[th]
\caption{Car detection (3D/BEV) performance of AVOD with \textit{latent ensemble layers (LEL)} against \textit{Gaussian} \ssn~and \asn~on the KITTI validation set.}
\label{table:result_car_lel_full_asn_ssn}
\begin{center}
\begin{scriptsize}
\begin{tabular}{lcccccc}
\toprule
	(Data) Train Algo. & Easy & Moderate & Hard & Easy & Moderate & Hard\\
\midrule
	(Clean Data) & \multicolumn{3}{c}{$\APTD (\%)$} & \multicolumn{3}{c}{$\APBEV (\%)$}\\ 
	\cmidrule(r){2-4} \cmidrule(r){5-7}
	AVOD \citep{ku2018joint} & \bf{77.79} & \bf{67.69} & \bf{66.31} & \bf{88.90} & \bf{85.64} & \bf{78.86}\\
	+\textsc{TrainASN} & 75.00 & 64.75 & 58.28 & 88.30 & 78.60 & 77.23\\
	+\textsc{TrainSSN} & 74.25 & 65.00 & 63.83 & 87.88 & 78.84 & 77.66\\
	+\textsc{TrainSSNAlt} & 76.04 & 66.42 & 64.41 & 88.80 & 79.53 & 78.53\\
\midrule
	(Gaussian ASN) & \multicolumn{3}{c}{$\APTD (\%)$} & \multicolumn{3}{c}{$\APBEV (\%)$}\\
	\cmidrule(r){2-4} \cmidrule(r){5-7}
	AVOD \citep{ku2018joint} & $46.79 {\pm} 0.37$ & $41.46 {\pm} 0.27$ & $36.31 {\pm} 0.20$ & $77.40 {\pm} 0.34$ & $67.46 {\pm} 0.11$ & $59.53 {\pm} 0.11$ \\
	+\textsc{TrainASN} & $\mathbf{74.24 {\pm} 0.29}$ & $\mathbf{63.47 {\pm} 0.18}$ & $\mathbf{57.25 {\pm} 0.19}$ & $87.72 {\pm} 0.12$ & $\mathbf{77.89 {\pm} 0.09}$ & $\mathbf{70.36 {\pm} 0.05}$\\
	+\textsc{TrainSSN} & $67.69 {\pm} 0.28$ & $55.74 {\pm} 0.30$ & $53.16 {\pm} 0.32$ & $\mathbf{87.73 {\pm} 0.16}$ & $77.80 {\pm} 0.15$ & $70.00 {\pm} 0.10$\\
	+\textsc{TrainSSNAlt} & $63.72 {\pm} 0.40$ & $53.15 {\pm} 0.29$ & $48.17 {\pm} 0.22$ & $85.36 {\pm} 0.08$ & $75.60 {\pm} 0.08$ & $69.17 {\pm} 0.03$ \\
\midrule
	(Gaussian SSN) & \multicolumn{3}{c}{$\min\APTD (\%)$} & \multicolumn{3}{c}{$\min\APBEV (\%)$}\\
	\cmidrule(r){2-4} \cmidrule(r){5-7}
	AVOD \citep{ku2018joint} & $61.97 {\pm} 0.55$ & $53.95 {\pm} 0.42$ & $47.24 {\pm} 0.27$ & $79.44 {\pm} 0.09$ & $72.46 {\pm} 3.14$ & $68.25 {\pm} 0.06$\\
	+\textsc{TrainASN} & $\mathbf{74.24 {\pm} 0.38}$ & $58.25 {\pm} 0.16$ & $\mathbf{56.13 {\pm} 0.10}$ & $88.10 {\pm} 0.26$ & $\mathbf{78.19 {\pm} 0.13}$ & $70.42 {\pm} 0.07$ \\
	+\textsc{TrainSSN} & $68.16 {\pm} 0.88$ & $\mathbf{60.39 {\pm} 0.38}$ & $56.04 {\pm} 0.28$ & $\mathbf{88.12 {\pm} 0.16}$ & $78.17 {\pm} 0.06$ & $70.21 {\pm} 0.05$\\
	+\textsc{TrainSSNAlt} & $68.63 {\pm} 0.40$ & $55.48 {\pm} 0.16$ & $54.42 {\pm} 0.17$
	& $86.51 {\pm} 0.46$ & $76.85 {\pm} 0.11$ & $\mathbf{71.95 {\pm} 2.72}$\\
	\midrule
	(Gaussian SSN) & \multicolumn{3}{c}{$\max \text{Diff}\APTD (\%)$} & \multicolumn{3}{c}{$\max \text{Diff}\APBEV (\%)$}\\
	AVOD \citep{ku2018joint} & $3.75 {\pm} 2.05$ & $0.98 {\pm} 0.55$ & $5.95 {\pm} 0.40$ & $7.28 {\pm} 0.37$ & $4.46 {\pm} 3.25$ & $\mathbf{1.25 {\pm} 0.13}$\\
	+\textsc{TrainASN} & $\mathbf{1.54 {\pm} 0.40}$ & $\mathbf{0.85 {\pm} 0.24}$ & $\mathbf{0.83 {\pm} 0.25}$ & $0.92 {\pm} 0.17$ & $1.09 {\pm} 0.14$ & $7.44 {\pm} 0.08$ \\
	+\textsc{TrainSSN} & $4.61 {\pm} 1.16$ & $2.51 {\pm} 0.50$ & $0.74 {\pm} 0.46$ & $\mathbf{0.16 {\pm} 0.32}$ & $\mathbf{0.72 {\pm} 0.14}$ & $7.10 {\pm} 0.14$\\
	+\textsc{TrainSSNAlt} & $4.65 {\pm} 1.04$ & $7.88 {\pm} 0.46$ & $2.90 {\pm} 0.45$ & $1.12 {\pm} 0.71$ & $1.83 {\pm} 0.17$ & $3.42 {\pm} 2.84$\\
	\bottomrule
\end{tabular}
\end{scriptsize}
\end{center}
\end{table}

\begin{table}[th]
	\caption{Car detection (3D/BEV) performance of AVOD with \textit{latent ensemble layers (LEL)} (trained with fine-tuning) against \textit{Gaussian} \ssn~and \asn~on the KITTI validation set.}
	\label{table:result_car_lel_rand_tune_asn_ssn}
	\begin{center}
		\begin{scriptsize}
			\begin{tabular}{lcccccc}
				\toprule
				(Data) Train Algo. & Easy & Moderate & Hard & Easy & Moderate & Hard\\
				\midrule
				(Clean Data) & \multicolumn{3}{c}{$\APTD (\%)$} & \multicolumn{3}{c}{$\APBEV (\%)$}\\ 
				\cmidrule(r){2-4} \cmidrule(r){5-7}
				AVOD \citep{ku2018joint} & \bf{77.79} & \bf{67.69} & \bf{66.31} & \bf{88.90} & \bf{85.64} & 78.86\\
				+\textsc{TrainASN} & 74.65 & 65.40 & 63.40 & 88.18 & 79.21 & 78.42\\
				+\textsc{TrainSSN} & 76.95 & 67.22 & 65.66 & 88.77 & 79.74 & \bf{78.96}\\
				+\textsc{TrainSSNAlt} & 76.81 & 67.46 & 66.12 & 88.47 & 79.62 & 78.86\\
				\midrule
				(Gaussian ASN) & \multicolumn{3}{c}{$\APTD (\%)$} & \multicolumn{3}{c}{$\APBEV (\%)$}\\
				\cmidrule(r){2-4} \cmidrule(r){5-7}
				AVOD \citep{ku2018joint} & $46.79 {\pm} 0.37$ & $41.46 {\pm} 0.27$ & $36.31 {\pm} 0.20$ & $77.40 {\pm} 0.34$ & $67.46 {\pm} 0.11$ & $59.53 {\pm} 0.11$ \\
				+\textsc{TrainASN} & $\mathbf{63.73 {\pm} 0.24}$ & $\mathbf{53.16 {\pm} 0.16}$ & $\mathbf{47.79 {\pm} 0.17}$ & $\mathbf{80.18 {\pm} 0.07}$ & $\mathbf{76.26 {\pm} 0.03}$ & $\mathbf{69.12 {\pm} 0.04}$\\
				+\textsc{TrainSSN} & $60.80 {\pm} 0.48$ & $47.73 {\pm} 0.13$ & $45.67 {\pm} 0.15$ & $79.82 {\pm} 0.22$ & $69.66 {\pm} 0.10$ & $68.38 {\pm} 0.10$\\
				+\textsc{TrainSSNAlt} & $52.25 {\pm} 1.47$ & $43.77 {\pm} 0.62$ & $37.91 {\pm} 0.48$ & $77.51 {\pm} 0.12$ & $67.32 {\pm} 0.09$ & $59.65 {\pm} 0.10$ \\
				\midrule
				(Gaussian SSN) & \multicolumn{3}{c}{$\min\APTD (\%)$} & \multicolumn{3}{c}{$\min\APBEV (\%)$}\\
				\cmidrule(r){2-4} \cmidrule(r){5-7}
				AVOD \citep{ku2018joint} & $61.97 {\pm} 0.55$ & $53.95 {\pm} 0.42$ & $47.24 {\pm} 0.27$ & $79.44 {\pm} 0.09$ & $72.46 {\pm} 3.14$ & $68.25 {\pm} 0.06$\\
				+\textsc{TrainASN} & $\mathbf{68.08 {\pm} 0.44}$ & $\mathbf{57.28 {\pm} 0.18}$ & $\mathbf{55.27 {\pm} 0.20}$ & $86.45 {\pm} 0.08$ & $77.19 {\pm} 0.08$ & $69.57 {\pm} 0.08$ \\
				+\textsc{TrainSSN} & $67.98 {\pm} 1.31$ & $55.61 {\pm} 0.23$ & $53.76 {\pm} 0.20$ & $\mathbf{86.87 {\pm} 0.12}$ & $\mathbf{77.56 {\pm} 0.05}$ & $\mathbf{69.81 {\pm} 0.08}$\\
				+\textsc{TrainSSNAlt} & $62.76 {\pm} 0.41$ & $52.14 {\pm} 0.26$ & $46.55 {\pm} 0.13$
				& $85.34 {\pm} 2.36$ & $75.72 {\pm} 0.04$ & $68.60 {\pm} 0.02$\\
				\bottomrule
			\end{tabular}
		\end{scriptsize}
	\end{center}
\end{table}

\paragraph{Concatenation} Our analyses in Section \ref{sec:sinrobustness} assume to use a linear fusion model with a simple concatenation strategy. Therefore, we first train the AVOD model with concatenation fusion layers on clean data and fine-tune with different training strategies. Interestingly, a simple data augmentation strategy \textsc{TrainSSNAlt} does not work well in this case, and \textsc{TrainASN} algorithm learns the best robust model. Unlike our simple linear model deep learning jointly learns both feature representation and weights for the fusion layers. Also, concatenated convolutional features have large number of channels which are mixed without sparse constraints. Therefore, this may lead to a model with too complex joint feature representation which needs stronger guideline in optimization steps.

\begin{table}[th]
	\caption{Car detection (3D/BEV) performance of AVOD with \textit{concatenation} fusion layers (trained with fine-tuning) against \textit{Gaussian} \ssn~and \asn~on the KITTI validation set.}
	\label{table:result_car_concat_rand_asn_ssn}
	\begin{center}
		\begin{scriptsize}
			\begin{tabular}{lcccccc}
				\toprule
				(Data) Train Algo. & Easy & Moderate & Hard & Easy & Moderate & Hard\\
				\midrule
				(Clean Data) & \multicolumn{3}{c}{$\APTD (\%)$} & \multicolumn{3}{c}{$\APBEV (\%)$}\\ 
				\cmidrule(r){2-4} \cmidrule(r){5-7}
				AVOD \citep{ku2018joint} & \bf{78.40} & \bf{74.88} & \bf{67.78} & \bf{89.74} & \bf{87.76} & \bf{79.83}\\
				+\textsc{TrainASN} & 72.89 & 63.47 & 62.22 & 88.44 & 84.97 & 78.88\\
				+\textsc{TrainSSN} & 76.15 & 66.79 & 65.78 & 89.02 & 86.06 & 79.29\\
				+\textsc{TrainSSNAlt} & 76.46 & 72.98 & 66.94 & 89.07 & 86.39 & 79.34\\
				\midrule
				(Gaussian ASN) & \multicolumn{3}{c}{$\APTD (\%)$} & \multicolumn{3}{c}{$\APBEV (\%)$}\\
				\cmidrule(r){2-4} \cmidrule(r){5-7}
				AVOD \citep{ku2018joint} & $16.50 {\pm} 2.27$ & $15.12 {\pm} 0.06$ & $15.06 {\pm} 0.08$ & $25.81 {\pm} 0.23$ & $25.38 {\pm} 0.18$ & $17.45 {\pm} 0.08$ \\
				+\textsc{TrainASN} & $\mathbf{69.21 {\pm} 0.24}$ & $\mathbf{54.85 {\pm} 0.08}$ & $\mathbf{53.30 {\pm} 0.08}$ & $\mathbf{86.07 {\pm} 0.11}$ & $\mathbf{76.42 {\pm} 0.04}$ & $\mathbf{69.54 {\pm} 0.02}$\\
				+\textsc{TrainSSN} & $62.05 {\pm} 0.36$ & $50.35 {\pm} 2.58$ & $46.04 {\pm} 0.25$ & $79.21 {\pm} 0.08$ & $69.31 {\pm} 0.10$ & $61.21 {\pm} 0.06$\\
				+\textsc{TrainSSNAlt} & $33.86 {\pm} 2.85$ & $27.99 {\pm} 0.64$ & $22.59 {\pm} 0.60$ & $42.65 {\pm} 0.18$ & $41.77 {\pm} 0.18$ & $34.13 {\pm} 0.12$ \\
				\midrule
				(Gaussian SSN) & \multicolumn{3}{c}{$\min\APTD (\%)$} & \multicolumn{3}{c}{$\min\APBEV (\%)$}\\
				\cmidrule(r){2-4} \cmidrule(r){5-7}
				AVOD \citep{ku2018joint} & $31.23 {\pm} 0.31$ & $30.27 {\pm} 0.13$ & $30.49 {\pm} 0.18$ & $43.04 {\pm} 0.16$ & $42.81 {\pm} 0.10$ & $42.96 {\pm} 0.08$\\
				+\textsc{TrainASN} & $\mathbf{68.21 {\pm} 0.37}$ & $54.50 {\pm} 0.26$ & $47.91 {\pm} 0.21$ & $\mathbf{86.66 {\pm} 0.11}$ & $\mathbf{76.95 {\pm} 0.11}$ & $\mathbf{69.70 {\pm} 0.08}$ \\
				+\textsc{TrainSSN} & $64.39 {\pm} 0.23$ & $\mathbf{55.12 {\pm} 0.21}$ & $\mathbf{48.38 {\pm} 0.14}$ & $79.71 {\pm} 0.07$ & $70.05 {\pm} 0.07$ & $69.32 {\pm} 0.10$\\
				+\textsc{TrainSSNAlt} & $44.25 {\pm} 0.49$ & $37.23 {\pm} 0.44$ & $37.58 {\pm} 0.34$ & $59.06 {\pm} 0.12$ & $51.19 {\pm} 0.08$ & $51.28 {\pm} 0.06$\\
				\bottomrule
			\end{tabular}
		\end{scriptsize}
	\end{center}
\end{table}

\paragraph{Results on downsampling corruption} Downsampling the LIDAR sensor is important as it is not clear whether a model trained with a high-resolution sensor will still work with a low-resolution one. In fact, reducing the number of lasers of a LIDAR is directly related to its price, which an important practical issue in deploying an actual autonomous vehicle. As the rotating LIDAR sensor used in the KITTI dataset outputs point clouds with a horizontal structure, an RGB image's horizontal lines are also set to black to match the information loss ratio 1/4. Table \ref{table:result_car_lel_full_asn_ssn_downsample} fully reports the performance of AVOD using our LEL when downsampling is considered as a corruption method.

\begin{table}[th]
	\caption{Car detection (3D/BEV) performance of AVOD with \textit{latent ensemble layers (LEL)} against \textit{downsampling} \ssn~and \asn~on the KITTI validation set.}
	\label{table:result_car_lel_full_asn_ssn_downsample}
	\begin{center}
		\begin{scriptsize}
			\begin{tabular}{lcccccc}
				\toprule
				(Data) Train Algo. & Easy & Moderate & Hard & Easy & Moderate & Hard\\
				\midrule
				(Clean Data) & \multicolumn{3}{c}{$\APTD (\%)$} & \multicolumn{3}{c}{$\APBEV (\%)$}\\ 
				\cmidrule(r){2-4} \cmidrule(r){5-7}
				AVOD \citep{ku2018joint} & \bf{77.79} & \bf{67.69} & \bf{66.31} & 88.90 & \bf{85.64} & \bf{78.86}\\
				+\textsc{TrainASN} & 71.74 & 61.78 & 60.26 & 87.29 & 77.08 & 75.89\\
				+\textsc{TrainSSN} & 75.54 & 66.26 & 63.72 & 88.07 & 79.18 & 78.03\\
				+\textsc{TrainSSNAlt} & 76.22 & 66.05 & 63.87 & \bf{89.00} & 79.65 & 78.03\\
				\midrule
				(Downsample ASN) & \multicolumn{3}{c}{$\APTD (\%)$} & \multicolumn{3}{c}{$\APBEV (\%)$}\\
				\cmidrule(r){2-4} \cmidrule(r){5-7}
				AVOD \citep{ku2018joint} & 36.13 & 27.39 & 26.39 & 77.60 & 59.84 & 51.82 \\
				+\textsc{TrainASN} & \bf{71.30} & \bf{56.04} & \bf{49.08} & 85.66 & \bf{70.17} & \bf{68.55}\\
				+\textsc{TrainSSN} & 64.88 & 48.92 & 47.06 & \bf{86.21} & 69.26 & 61.48\\
				+\textsc{TrainSSNAlt} & 48.98 & 36.30 & 31.06 & 75.00 & 51.35 & 49.60 \\
				\midrule
				(Downsample SSN) & \multicolumn{3}{c}{$\min\APTD (\%)$} & \multicolumn{3}{c}{$\min\APBEV (\%)$}\\
				\cmidrule(r){2-4} \cmidrule(r){5-7}
				AVOD \citep{ku2018joint} & 61.70 & 51.66 & 46.17 & 86.08 & 69.99 & 61.55\\
				+\textsc{TrainASN} & 65.74 & 53.49 & 51.35 & 82.27 & 67.88 & 65.79\\
				+\textsc{TrainSSN} & \bf{73.33} & \bf{57.85} & \bf{54.91} & \bf{86.61} & \bf{76.07} & \bf{68.59}\\
				+\textsc{TrainSSNAlt} & 64.77 & 53.34 & 48.29 & 85.27 & 69.87 & 67.77\\
				\midrule
				(Downsample SSN) & \multicolumn{3}{c}{$\max \text{Diff}\APTD (\%)$} & \multicolumn{3}{c}{$\max \text{Diff}\APBEV (\%)$}\\
				AVOD \citep{ku2018joint} & 11.71 & 5.88 & 3.59 & 1.96 & 7.60 & 8.65\\
				+\textsc{TrainASN} & 10.00 & 11.34 & 11.76 & 6.53 & 11.23 & 12.40\\
				+\textsc{TrainSSN} & \bf{0.94} & 5.71 & 3.11 & 1.74 & 2.36 & 9.00\\
				+\textsc{TrainSSNAlt} & 6.98 & \bf{3.63} & \bf{1.34} & \bf{1.67} & \bf{0.12} & \bf{0.81}\\
				\bottomrule
			\end{tabular}
		\end{scriptsize}
	\end{center}
\end{table}

\end{document}